\documentclass[journal]{IEEEtran}
\usepackage{amsmath,amsfonts}
\usepackage{algorithm}
\usepackage{array}
\usepackage[caption=false,font=normalsize,labelfont=sf,textfont=sf]{subfig}
\usepackage{textcomp}
\usepackage{stfloats}
\usepackage{url}
\usepackage{verbatim}
\usepackage{graphicx}
\usepackage{cite}
\usepackage{graphicx}
\usepackage{subcaption}
\usepackage{siunitx}
\usepackage{caption}
\usepackage{algpseudocode}    
\usepackage{color}
\usepackage[table]{xcolor}

\usepackage{booktabs}
\usepackage{multirow}
\usepackage{makecell}
\usepackage{rotating}
\usepackage{bbding}
\usepackage{threeparttable}

\usepackage[colorlinks,bookmarksopen,bookmarksnumbered,citecolor=blue, linkcolor=red, urlcolor=magenta]{hyperref}
\usepackage[switch]{lineno} 

\begin{document}

\title{A Multi-Task Targeted Learning Framework \\ for Lithium-Ion Battery State-of-Health and \\ Remaining Useful Life}

\author{Chenhan Wang, Zhengyi Bao,~\IEEEmembership{Member,~IEEE}, Huipin Lin,~\IEEEmembership{Member,~IEEE},\\
Jiahao Nie,~\IEEEmembership{Member,~IEEE}, Chunxiang Zhu

\thanks{This work was supported by the Zhejiang Provincial Major Research and Development Project of China under Grant 2025C01004 and 2024C01011, the Provincial Public University Fundamental Research Fund under Grant GK249909299001-302, the Wenzhou Municipal Basic Scientific Research Project under grant G20240013, the Baima Lake Laboratory Joint Fund of Zhejiang Provincial Natural Science Foundation of China under Grant No. LBMHY25B030001, and Zhejiang Key Laboratory of Intelligent Vehicle Electronics Research.
(\textit{Corresponding author: Zhengyi Bao, Jiahao Nie})} 
\thanks{Chenhan Wang, Zhengyi Bao and Huipin Lin are with the School of Electronics and Information Engineering, Hangzhou Dianzi University, Hangzhou 310018, China (e-mail: 22080626@hdu.edu.cn; baozy@hdu.edu.cn, linhuipin@hdu.edu.cn.}
\thanks{Jiahao Nie is with the School of Information Technology and Artificial Intelligence, Zhejiang University of Finance and Economics, Hangzhou 310018, China (e-mail: jhnie@zufe.edu.cn).}
\thanks{Chunxiang Zhu is with the Engineering Training Center, China Jiliang University, Hangzhou 310018, China (e-mail: 16a2900069@cjlu.edu.cn).}}

\maketitle

\begin{abstract}
Accurately predicting the state-of-health (SOH) and remaining useful life (RUL) of lithium-ion batteries is crucial for ensuring the safe and efficient operation of electric vehicles while minimizing associated risks. However, current deep learning methods are limited in their ability to selectively extract features and model time dependencies for these two parameters. Moreover, most existing methods rely on traditional recurrent neural networks, which have inherent shortcomings in long-term time-series modeling. To address these issues, this paper proposes a multi-task targeted learning framework for SOH and RUL prediction, which integrates multiple neural networks, including a multi-scale feature extraction module, an improved extended LSTM, and a dual-stream attention module. First, a feature extraction module with multi-scale CNNs is designed to capture detailed local battery decline patterns. Secondly, an improved extended LSTM network is employed to enhance the model's ability to retain long-term temporal information, thus improving temporal relationship modeling. Building on this, dual-stream attention module—comprising polarized attention and sparse attention are introduced to selectively focus on key information relevant to SOH and RUL, respectively, by assigning higher weights to important features. Finally, a many-to-two mapping is achieved through the dual-task layer. To optimize the model's performance and reduce the need for manual hyperparameter tuning, the Hyperopt optimization algorithm is used. Extensive comparative experiments on battery aging datasets demonstrate that the proposed method reduces the average RMSE for SOH and RUL predictions by 111.3\% and 33.0\%, respectively, compared to traditional and state-of-the-art methods. The code will be made publicly available at: \url{https://github.com/wch1121/Joint-prediction-of-SOH-and-RUL}.
\end{abstract}


\section{Introduction}
\label{introduction}

\IEEEPARstart{W}{ith} the rapid development of electric vehicles (EVs), lithium-ion batteries have become widely adopted due to their high energy density and low self-discharge rate \cite{r1_3}. However, as the number of charge-discharge cycles increases, battery performance inevitably degrades, leading to potential safety hazards \cite{1}. To ensure the safe and stable operation of power batteries, it is essential to accurately assess their status. Various indicators within the battery management system (BMS) are employed to evaluate battery health \cite{2}. Among these, the state-of-health (SOH) and remaining useful life (RUL) are the most critical performance indicators. Accurate prediction of SOH and RUL is vital for optimizing battery management strategies and ensuring operational safety \cite{4}. SOH is commonly expressed as $\mathrm{SOH} = \frac{{{C_{current}}}}{{{C_{rated}}}} \times 100\% $, where ${C_{current}}$ denotes the current capacity of the battery, and ${C_{rated}}$ represents its rated capacity. RUL is typically defined as $\mathrm{RUL} = {n_{EOL}} - {n_{current}}$, where ${n_{EOL}}$ is the total number of cycles at which the battery reaches its end-of-life (EOL), and ${n_{current}}$ is the current number of cycles completed.

Numerous researchers have focused on the prediction of SOH and RUL, with existing methods broadly categorized into two types: model-based methods and data-driven methods. Model-based methods rely on equivalent circuit models (ECMs) and electrochemical models (EMs), often coupled with filtering algorithms, to make predictions. ECMs simulate battery degradation using electrical circuit components \cite{7}. \cite{8} proposed a temperature-controlled second-order R-CPE ECM to capture the non-ideal capacitance characteristics of the electrode surfaces for further accurate understanding of the chemical reactions during battery aging. However, these models are highly sensitive to external factors and heavily dependent on model parameters, making them unsuitable for complex, real-world applications. EMs simulate degradation based on the battery’s electrochemical properties \cite{9}. \cite{10}. Although these methods have clear physical significance, their parameters are often complex and diverse, limiting their generalizability. In addition, \cite{r3_4} proposed a battery life prediction method based on physical modeling, which simulates the full lifecycle evolution of batteries through adaptive extrapolation. However, such models generally lack generalizability and are not well-suited for tracking the degradation of diverse battery types under real-world operating conditions. Overall, model-based methods depend heavily on the accuracy of battery modeling, but existing models struggle to capture the inherently complex aging mechanisms of batteries. Additionally, updating or reconstructing these models to account for different battery types and aging stages remains a significant challenge \cite{11}.

With the rapid advancement of machine learning (ML), data-driven models have become widely used for monitoring and predicting battery status. These methods work by constructing a learning model that links externally observable battery parameters to battery degradation, enabling the model to perform nonlinear many-to-one mapping \cite{12}. Data-driven models can be broadly divided into two categories: traditional ML algorithms and artificial neural network (NN) algorithms. Traditional ML algorithms include Support Vector Machine \cite{13}, Relevance Vector Machine, and Random Forest. \cite{r3_1} used Gaussian process (GP) regression to realize short-term and long-term battery life prediction. \cite{r3_5} proposed feature-based and Extreme Gradient Boosting (XGBoost) methods for RUL prediction, which require relatively less raw data for effective performance. \cite{18} used variable forgetting factor online continuous limit learning machine to predict the SOH of the battery, and particle filtering algorithm to predict the RUL. While these methods are relatively simple to implement, they exhibit limited scalability and generalizability. In recent years, artificial NNs have gained increasing attention due to their adaptability and versatility \cite{19}. Convolutional Neural Networks (CNNs) excel at feature extraction \cite{nie1, nie2}; however, they inherently lack the ability to effectively capture temporal dependencies in time series data. In contrast, Recurrent Neural Networks (RNNs) and their advanced variants, such as Gated Recurrent Units (GRUs) and Long Short-Term Memory (LSTM) networks \cite{r1_4}, are well-suited for modeling sequential patterns due to their strong temporal learning capabilities \cite{21}. When integrated with CNNs, these recurrent architectures can leverage the complementary strengths of both network types, enhancing performance in time series prediction tasks. \cite{r1_1} proposed a SOH prediction network based on the CNN-Multi-gate Mixture of GRU, while \cite{r1_2} introduced a CNN-enhanced feature combination-bidirectional LSTM for SOH prediction. Both methods demonstrate that fusion models offer improved generalization capabilities. However, RNNs have inherent limitations, such as a small memory size and limited capacity, which hinders their ability to accurately predict long-term battery degradation, leading to reduced prediction accuracy. Recently, battery state prediction methods based on physical models have gained increasing attention \cite{r3_3}. These methods integrate physical principles into NN architectures to reduce reliance on large volumes of data. However, their performance is heavily dependent on the accuracy of the underlying physical models, which can, in turn, constrain the scalability and adaptability of the overall framework. Therefore, it is necessary to enhance the scalability and performance of the RNN-based architectures in processing time-series data, and to integrate efficient feature extraction and dynamic assignment networks to effectively address the predictive requirements of battery SOH and RUL.

Furthermore, in battery degradation, SOH and RUL are two key indicators that exhibit a complex and interdependent relationship. While numerous studies have sought to jointly evaluate these parameters, the majority adopt a two-step prediction framework: initially, a NN is employed to estimate the battery’s SOH, and subsequently, the predicted SOH is used as input for a second network to estimate the RUL \cite{joule}. However, these sequential methods present several notable limitations. (1) The degradation trajectory of batteries varies under different usage scenarios, such as second-life applications or resale conditions. As a result, inferring RUL solely from SOH predictions may lead to inaccurate outcomes due to mismatched degradation patterns. (2) The two-step method propagates errors—any inaccuracies in the initial SOH prediction significantly impact the downstream RUL estimation, thereby compromising overall prediction accuracy. (3) Similar to single-parameter prediction models, this two-step framework fails to reduce computational complexity or improve efficiency.

To address these challenges, this paper proposes a multi-task joint prediction network that enables end-to-end online estimation of both SOH and RUL simultaneously. The main contributions of this study are as follows:

\begin{itemize}
    \item This paper introduces a unified NN framework for the joint estimation of SOH and RUL, enabling simultaneous prediction of both parameters. In our previous work, correlation analysis revealed strong associations between raw battery voltage sequences and the corresponding SOH and RUL. Given the instability of current profiles under real-world EV operations and the difficulty of manual feature extraction, this paper utilizes raw voltage sequences as input to model a many-to-two mapping relationship directly.
    \item This paper proposes an improved extended LSTM network for relational modeling of battery time-series information, incorporating exponential gating and a novel hybrid data storage technique. These additions enable the network to adaptively adjust storage decisions based on the characteristic changes in battery degradation over time.
    \item To address the distinct output requirements for SOH and RUL, specialized attention mechanisms are developed to facilitate targeted learning of critical information.
    \item To streamline the process and minimize manual intervention, the Hyperopt optimizer is utilized to automatically identify the optimal network hyperparameters.
\end{itemize}

\section{Methodologies}

\begin{figure*}[t]
    \centering
    \includegraphics[width=0.92\linewidth]{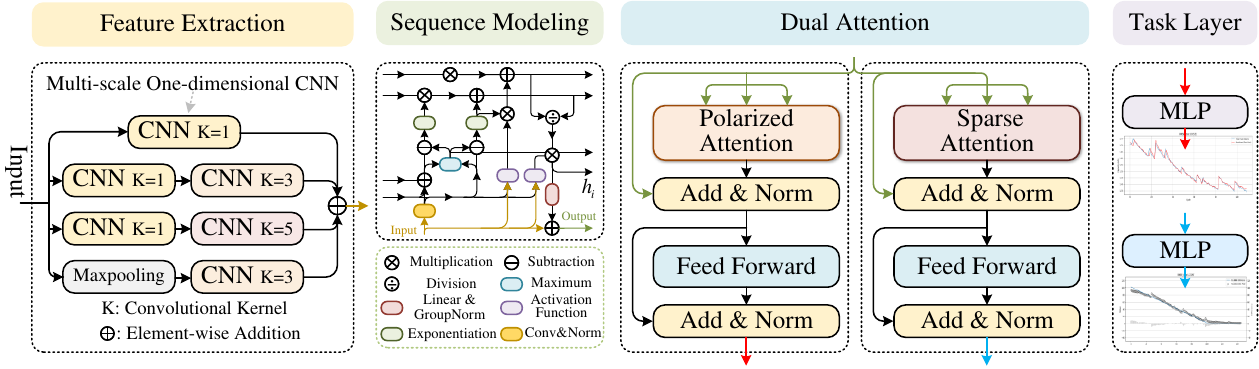}
    \caption{Flowchart of the proposed NN based on the fusion of FEM, IE-LSTM and DSAM.}
    \label{fig1}
    \vspace{-7.5pt}
\end{figure*}

The overall network framework is illustrated in Fig. \ref{fig1}. It comprises four key components: a feature extraction module (FEM), an improved LSTM module (IE-LSTM), a dual-stream attention module (DSAM), and a task layer. Each component is detailed in the following sections.

\label{method}

\subsection{Feature extraction module}

To extract significant features from the original battery input data for subsequent network modeling, a multi-scale FEM is initially constructed. This module consists of multiple layers of one-dimensional CNNs operating at different scales, incorporating the activation function to introduce nonlinearity and enhance the overall expressive capacity of the network. Additionally, this module employs sparse matrix decomposition to facilitate dense matrix computations, thereby increasing convergence speed. In this paper, the input matrix for the network is denoted as ${X_i} = {\left[ {{V_1},{V_2},...,{V_L}} \right]^T},{X_i} \in {\mathbb{R}^{L \times 1}}$, where $L$ represents the length of the input sequence and ${V_i}$ denotes the voltage value at the corresponding time point during the $i$-th charge/discharge cycle. The mathematical formulation of the module is provided as follows:
\begin{equation}
X_i^{FEM} \Rightarrow {F_{FEM}}\left( {{X_i}} \right) = concat\left( {X_i^{B{r_1}},X_i^{B{r_2}},X_i^{B{r_3}},X_i^{B{r_4}}} \right)
\end{equation}
where $X_i^{B{r_j}}$ (ranging from 1 to 4) represents a branch comprising convolutional layers of varying scales, while $concat( \cdot )$ denotes the concatenation along a specified dimension. The feature matrix resulting from the FEM is represented as $X_i^{FEM} \in {\mathbb{R}^{L \times C}}$, with $c$ indicating the number of feature channels.  Given that the convolutional layer with a kernel size of 5 introduced in the third branch $X_i^{B{r_3}}$ incurs significant computational overhead, we opt to incorporate a convolutional kernel with a size of 1 in branches $X_i^{B{r_2}}$, $X_i^{B{r_3}}$, and $X_i^{B{r_4}}$ to perform dimensionality reduction. This adjustment reduces computational demands while enabling a greater number of convolutional kernels to be stacked within the same sensory field, thereby allowing the module to extract a richer set of features. The specific formulations for each branch within the FEM are outlined below:
\begin{equation}
{X_i^{B{r_1}} = Con{v_1}\left( X \right)}
\end{equation}
\begin{equation}
{X_i^{B{r_2}} = Con{v_2}\left[ {Con{v_1}\left( X \right)} \right]}
\end{equation}
\begin{equation}
{X_i^{B{r_3}} = Con{v_4}\left[ {Con{v_3}\left( X \right)} \right]}
\end{equation}
\begin{equation}
{X_i^{B{r_4}} = Con{v_5}\left[ {MaxPool\left( X \right)} \right]}
\end{equation}
where $Con{v_j}\left(  \cdot  \right),j = [1-5]$ denotes a convolutional layer cascading an activation function. $MaxPool\left(  \cdot  \right)$ denotes a max pooling layer.

\subsection{Improved extended LSTM}

Sensor reading errors and the intricate mechanisms governing battery behavior can result in fluctuations and variations in the battery's state. Moreover, degradation data is typically collected over extended periods, with informative features concentrated within specific temporal segments. These characteristics impose higher demands on the modeling and memory capacity of predictive networks. Traditional LSTM networks rely on a fixed sigmoid gating mechanism, which limits their ability to adapt efficiently to rapid feature variations or noise within feature-dense sequence segments. To overcome this limitation, we propose the IE-LSTM as the core component for modeling battery sequence data. By incorporating an exponential activation gating function, the IE-LSTM enhances the responsiveness of the input and forget gates to dynamic changes in the input. This design enables the model to effectively amplify critical feature signals within voltage trajectories while suppressing noise interference, thereby improving robustness and predictive accuracy. The specific computational process of this model can be expressed as follows:

\begin{equation}
X_i^{IE - LSTM} \Rightarrow {F_{IE - LSTM}}\left( {X_i^{FEM}} \right) + X_i^{FEM}
\end{equation}
where ${F_{IE - LSTM}}\left(  \cdot  \right)$ denotes the overall function of the IE-LSTM, while $X_i^{IE - LSTM}$ signifies the output that is processed through the module. The specific computational flow of the IE-LSTM can be described as follows:
\begin{equation}
{F_{IE - LSTM}}\left( {X_i^{FEM}} \right) = L{n^1}\left\{ {L{n^2}\left( {{H_i^{sL}}} \right) \odot \vartheta \left[ {L{n^3}\left( {{H_i^{sL}}} \right)} \right]} \right\}
\end{equation}
$L{n^i}\left(  \cdot  \right),i = [1,2,3]$ refers to three fully connected layers with varying neuron sizes, $\vartheta \left(  \cdot  \right)$ denotes the GeLU, $ \odot $ indicates the element-wise multiplication operation, and $H_i^{sL}$ represents the output data processed by the IE-LSTM structure, as depicted in Fig. \ref{fig2}. Unlike traditional LSTM, the inputs of each cell $x_i^{FEM}$ in the IE-LSTM are first processed through a normalization layer and a convolutional layer to handle the original features prior to entering the forgetting gate and input gate. This approach aims to mitigate numerical stability issues and stabilize fluctuating inputs, where $x_i^{FEM}$ represents the feature vector within the feature matrix $X_i^{FEM}$.

\begin{equation}
{H_i^{sL} \Rightarrow {F_g}\left[ {\tilde X_i^{FEM},X_i^{FEM}} \right]}
\end{equation}

\begin{equation}
{\tilde X_i^{FEM} = Conv\left[ {Norm\left( {X_i^{FEM}} \right)} \right]}
\end{equation}
${F_g}\left(  \cdot  \right)$ represents the gating operation of the IE-LSTM, while $Norm\left(  \cdot  \right)$ denotes the normalization operation. Each vector within matrices $\tilde X_i^{FEM} = {\left[ {\tilde x_1^{FEM},\tilde x_2^{FEM},...,\tilde x_L^{FEM}} \right]^T}$ and $X_i^{FEM} = {\left[ {x_1^{FEM},x_2^{FEM},...,x_L^{FEM}} \right]^T}$ sequentially enters each gating unit of the IE-LSTM as follows:

\begin{equation}
\begin{array}{l}
{i_t} = \exp \left( {{{\tilde i}_t} - {m_t}} \right),{{\tilde i}_t} = W{h_{t - 1}} + W\tilde x_t^{FEM} + b\\
{f_t} = \exp \left( {{{\tilde f}_t} + {m_{t - 1}} - {m_t}} \right),{{\tilde f}_t} = W{h_{t - 1}} + W\tilde x_t^{FEM} + b\\
{m_t} = \max ({{\tilde f}_t} + {m_{t - 1}},{{\tilde i}_t}),{n_t} = {f_t}{n_{t - 1}} + {i_t}\\
{c_t} = {f_t}{c_{t - 1}} + {i_t}{z_t},{z_t} = \tanh \left( {W{h_{t - 1}} + Wx_t^{FEM} + b} \right)\\
{h_t} = {o_t}{c_t}/{n_t},{o_t} = \sigma \left( {W{h_{t - 1}} + Wx_t^{FEM} + b} \right)
\end{array}
\end{equation}
where $t$ takes values in the range of [1,L], $W$ represents the corresponding transition matrix weights, and $b$ denotes the associated bias matrix. $\mathrm{max}\left( { \cdot , \cdot } \right)$ refers to the operation of taking the maximum value from the input data, while $\exp \left(  \cdot  \right)$ , $\sigma \left(  \cdot  \right)$ and $\tanh \left(  \cdot  \right)$, correspond to the exponential, sigmoid, and tanh activation functions, respectively. Additionally, ${m_t}$ and ${n_t}$ are new state variables introduced into the IE-LSTM to stabilize and normalize the data, respectively. This step facilitates more efficient updates to the internal state of each cell in the IE-LSTM and allows for targeted modifications of stored information, enabling the cell to control the flow of information with greater flexibility.

\subsection{Dual-stream attention module}

To effectively learn critical information between input features, SOH and RUL, and to assign appropriate weight values to each, two separate streams with distinct attention mechanisms are constructed. Both streams within the DSAM employ the Transformer encoder architecture, incorporating elements such as residual connections and normalization operations to improve network stability, training efficiency, and the model's expressive capability.

\subsubsection{Polarized attention}

To accommodate complex operating conditions and better adapt to the aging trends within extensive data, a polarized attention mechanism is applied to the SOH prediction task \cite{30}. This mechanism enhances the model's ability to identify critical feature regions in voltage curves by integrating both channel polarization and spatial polarization strategies. Specifically, the channel polarization branch applies global average pooling in combination with a gating mechanism to produce channel weights that approximate binary values. This design forces the model to concentrate on differential voltage features that are strongly correlated with capacity degradation, while simultaneously suppressing noise fluctuations and other irrelevant factors. In parallel, the spatial polarization branch employs dynamic threshold binarization to directly locate key segments within charge–discharge profiles (e.g., voltage drop points at the end of the constant-voltage phase), which are often highly indicative of the final SOH. The mathematical representation is as follows:

\begin{equation}
\begin{array}{c}
X_i^{DSAM - SOH} \Rightarrow F_{DSAM}^{\left( {SOH} \right)}\left( {X_i^{IE - LSTM}} \right)\\
 = Norm\left[ {{F_{ff}}\left( {X_i^{PA}} \right) + X_i^{PA}} \right]
\end{array}
\end{equation}
\begin{equation}
X_i^{PA} \Rightarrow Norm\left[ {{F_{PA}}\left( {X_i^{IE - LSTM}} \right) + X_i^{IE - LSTM}} \right]
\end{equation}
$X_i^{DSAM - SOH}$ represents the output, $F_{DSAM}^{\left( {SOH} \right)}\left(  \cdot  \right)$ denotes the attention mechanism within the SOH prediction stream, ${F_{PA}}\left(  \cdot  \right)$ indicates the polarized attention, and ${F_{ff}}\left(  \cdot  \right) = Ln\left\{ {\sigma \left[ {Ln\left(  \cdot  \right)} \right]} \right\}$ represents the feedforward neural network. The polarized attention mechanism specifically incorporates two branches—the channel branch and the spatial branch—linked in series. Within each branch, feature degradation and self-attention are applied to derive the attention weights, facilitating effective long-range modeling. The detailed operational steps are as follows:
\begin{equation}
{F_{PA}}\left( {X_i^{IE - LSTM}} \right) = X_i^{sh} \odot \left( {X_i^{ch} \odot X_i^{IE - LSTM}} \right)
\end{equation}
\begin{equation}
\begin{array}{c}
X_i^{ch} \Rightarrow {F_{ch}}\left( {X_i^{IE - LSTM}} \right)\\
 = Ln\left[ {{C_z}\left( {{C_v}\left( {X_i^{IE - LSTM}} \right) \otimes s\left( {{C_q}{{\left( {X_i^{IE - LSTM}} \right)}^r}} \right)} \right)} \right]
\end{array}
\end{equation}
\begin{equation}
\begin{array}{c}
X_i^{sh} \Rightarrow {F_{sh}}\left( {X_i^{ch}} \right)\\
 = {\left[ {{C_v}\left( {X_i^{ch}} \right) \otimes s\left( {MaxPool{{\left( {{C_q}\left( {X_i^{ch}} \right)} \right)}^r}} \right)} \right]^r}
\end{array}
\end{equation}
${C_q}$, ${C_v}$ and ${C_z}$ represent three CNNs, each with a kernel size of 3. ${\left(  \cdot  \right)^r}$ denotes the exchange of the last two dimensions of data, $s\left(  \cdot  \right)$ indicates the softmax activation function, and $ \otimes $ represents matrix multiplication.

\begin{figure*}[t]
    \centering
    \includegraphics[width=0.92\linewidth]{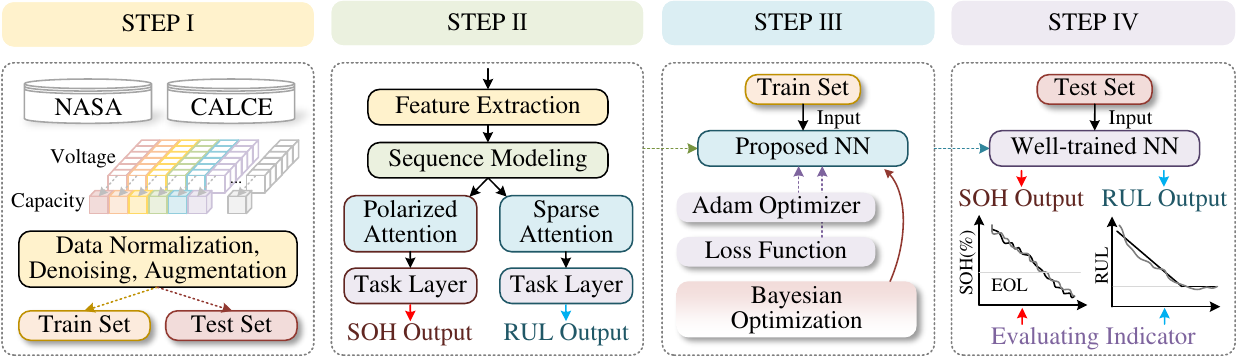}
    \caption{The proposed overall framework for SOH and RUL prediction.}
    \label{fig2}
    \vspace{-7.5pt}
\end{figure*}

\subsubsection{Sparse attention}

To reduce computational complexity, prioritize essential historical information related to RUL, and effectively capture temporal dependencies, sparse attention \cite{31} is employed as the attention mechanism within the RUL prediction stream. Unlike conventional attention mechanisms, which compute correlations across all sampling points, sparse attention dynamically learns to retain only the connections between each time step and its most relevant positions. In this way, it preserves critical features within the voltage curve while substantially lowering model complexity. The data representation is as follows:

\begin{equation}
\begin{array}{c}
X_i^{DSAM - RUL} \Rightarrow F_{DSAM}^{\left( {RUL} \right)}\left( {X_i^{IE - LSTM}} \right)\\
 = Norm\left[ {{F_{ff}}\left( {X_i^{SA}} \right) + X_i^{SA}} \right]
\end{array}
\end{equation}
\begin{equation}
X_i^{SA} \Rightarrow Norm\left[ {{F_{SA}}\left( {X_i^{IE - LSTM}} \right) + X_i^{IE - LSTM}} \right]
\end{equation}
where $X_i^{DSAM - RUL}$ denotes the output, $F_{DSAM}^{\left( {RUL} \right)}\left(  \cdot  \right)$ represents the attention mechanism within the RUL prediction stream, and ${F_{SA}}\left(  \cdot  \right)$ signifies the sparse attention. The sparse attention enhances prediction efficiency by reducing the computation required. It focuses only on representative attention weights, omitting dot products with minimal impact on the outcome. The mathematical expression for this mechanism is as follows:
\begin{equation}
{{F_{SA}}\left( {X_i^{IE - LSTM}} \right) = Pad\left[ {{\mathop{\rm softmax}\nolimits} \left( {\frac{{\bar Q{K^T}}}{{\sqrt d }}} \right)V,mean\left( V \right)} \right]}
\end{equation}
\begin{equation}
{Q = {W_Q}X_i^{IE - LSTM} = {{\left[ {{q_1},{q_2},...,{q_L}} \right]}^T}}
\end{equation}
\begin{equation}
{K = {W_K}X_i^{IE - LSTM} = {{\left[ {{k_1},{k_2},...,{k_L}} \right]}^T}}
\end{equation}
\begin{equation}
{V = {W_V}X_i^{IE - LSTM} = {{\left[ {{v_1},{v_2},...,{v_L}} \right]}^T}}
\end{equation}
where $\bar Q \in {\mathbb{R}^{U \times C}}$ is a sparse matrix about $Q$, $U = c * \ln L$. $c$ is an optional hyperparameter. $\bar Q$ contains only the $U$ query with the highest sparsity measure $\bar M\left( {{q_i},{K_S}} \right)$, and the expression for $\bar M\left( {{q_i},{K_S}} \right)$ is as follows:
\begin{equation}
\bar M\left( {{q_i},{K_S}} \right) = \max \left( {\frac{{{q_i}{K_S}}}{{\sqrt d }}} \right) - \frac{1}{L}\sum\limits_{j = 1}^S {\frac{{{q_i}\bar k_j^T}}{{\sqrt d }}}
\end{equation}
where ${K_S}$ denotes $K$ after randomly sampling $S$ value, i.e., ${K_S} = {\left[ {{{\bar k}_1},{{\bar k}_2},...,{{\bar k}_S}} \right]^T},{K_S} \in {^{S \times C}}$, where $S = L\ln L$. $\left[ {{\mathop{\rm softmax}\nolimits} \left( {\frac{{\bar Q{K^T}}}{{\sqrt d }}} \right)V} \right] \in {\mathbb{R}^{U \times C}}$, $Pad\left(  \cdot  \right)$ denotes feature filling, expanding the data dimension from $U \times C$ to $L \times C$, and $mean\left(  \cdot  \right)$ represents averaging.

\subsection{Task layer}
The two feature matrices resulting from the DSAM processing contain critical features highly correlated with SOH and RUL, respectively. To achieve nonlinear mapping between inputs and outputs and to shape the final output to provide two targeted outputs, a task layer is constructed following the attention layer.
\begin{equation}
\begin{array}{c}
X_i^{\left( {SOH} \right)} \Rightarrow F_{MLP}^{\left( {SOH} \right)}\left( {X_i^{DSAM - SOH}} \right)\\
 = Ln\left\{ {\delta \left[ {Ln\left( {X_i^{DSAM - SOH}} \right)} \right]} \right\}
\end{array}
\end{equation}
\begin{equation}
\begin{array}{c}
X_i^{\left( {RUL} \right)} \Rightarrow F_{MLP}^{\left( {RUL} \right)}\left( {X_i^{DSAM - RUL}} \right)\\
 = Ln\left\{ {\delta \left[ {Ln\left( {X_i^{DSAM - RUL}} \right)} \right]} \right\}
\end{array}
\end{equation}
where $F_{MLP}^{\left( {SOH} \right)}\left(  \cdot  \right)$ and $F_{MLP}^{\left( {RUL} \right)}\left(  \cdot  \right)$ represent the tasks for SOH and RUL prediction, respectively. The dimensionality reduction operation of the data, which facilitates the mapping to the desired output, is accomplished through the use of two fully connected layers followed by an activation layer.

\section{SOH and RUL prediction}
\label{3}

\subsection{Experimental dataset}

\begin{table}[ht]
\centering
\caption{Detailed description of the battery used in the experiments.}
\begin{tabular}{lccccc}
\toprule
 & \textbf{NASA} & \textbf{CALCE} & \textbf{XJTU} & \textbf{Oxford} & \textbf{MIT} \\
\midrule
Cell Chemistry & LCO & LCO & NCM & LCO & LFP \\
Rated Capacity & 2 Ah & 1.1 Ah & 2 Ah & 0.74 Ah & 1.1 Ah \\
EOL & 1.4 Ah & 0.77 Ah & 1.6 Ah & 0.59 Ah & 0.88 Ah \\
Saturation Voltage & 4.2 V & 4.2 V & 1.2 V & 4.2 V & 3.6 V \\
Battery Number & 28 & 15 & 55 & 8 & 124 \\
\bottomrule
\label{dataset_table}
\end{tabular}
\end{table}

This paper employs five battery aging datasets: the NASA dataset \cite{NASA} (available at URL \url{https://data.nasa.gov/download/brfb-gzcv/application%2F.zip}), the CALCE dataset \cite{CALCE} (available at URL \url{https://calce.umd.edu/battery-data}), the XJTU dataset \cite{XJTU} (available at URL \url{https://github.com/wang-fujin/PINN4SOH}), the Oxford dataset \cite{Oxford} (available at URL \url{https://ora.ox.ac.uk/objects/uuid:03ba4b01-cfed-46d3-9b1a-7d4a7bdf6fac}), and the MIT dataset \cite{MIT} (available at URL \url{https://data.matr.io/1/}). The EOL of a battery is typically defined as 70–80\% of its nominal rated capacity \cite{r3_6}. For the five datasets used in this paper, the specific EOL criteria are determined according to the documentation provided with the datasets and relevant published literature \cite{R22}, as summarized in Table \ref{dataset_table}. Comprehensive descriptions of each dataset are presented in Supplementary Material \uppercase\expandafter{\romannumeral1}.

\subsection{Procedure of SOH and RUL prediction}

The joint prediction process for SOH and RUL is illustrated in Fig. \ref{fig2} and comprises four distinct steps.

\noindent\textbf{Step A:} Data Preprocessing. To ensure consistency in input dimensions across different charge–discharge cycles, the raw data are preprocessed prior to network training. Since the lengths of charge/discharge sequences vary among cycles, the NASA dataset is interpolated to 4,000 points and the CALCE dataset to 1,200 points, with the length after interpolation determined by the maximum sequence length observed over the entire lifecycle. This strategy facilitates temporal alignment and enables efficient model training. To minimize distortion and information loss, interpolation is performed according to the following rules: (1) Determining interpolation length: the number of interpolated points is defined as the difference between the actual sequence length of a given cycle and the target length; (2) Position allocation: interpolation points are uniformly distributed across the sequence; (3) Value estimation: the value at each interpolated point is computed as the average of its two adjacent original points.

Except for Section \ref{Non-full}, all experiments are conducted using the complete charge–discharge data. The training/testing partition strategy and the selection of observation periods are summarized in Table \ref{Data_partitioning}.
According to the relevant literature \cite{c1}, the B0007 in the NASA dataset did not reach its defined EOL, as its minimum capacity was 1.4005 Ah—slightly above the EOL threshold of 1.4 Ah specified for NASA dataset. Consequently, in conjunction with the RUL prediction formula outlined in Section \ref{introduction}, subsequent RUL estimation is not feasible, and only SOH prediction is conducted for this case. A similar situation arises with the 50\_CH32 in the MIT dataset, whose minimum capacity of 0.8809 Ah remains marginally above the EOL threshold of 0.88 Ah defined for MIT batteries.

\begin{table}[t]
\centering
\caption{Data partitioning strategy, namely the division of training set/test set and the selection of observation cycle. OC represents observation cycle selection.}
\begin{tabular}{@{}>{\centering\arraybackslash}p{1.2cm}
                >{\centering\arraybackslash}p{1.0cm}
                >{\centering\arraybackslash}p{3.6cm}
                >{\centering\arraybackslash}p{1.3cm}@{}}
\toprule[0.5mm]
\multicolumn{2}{c}{Dataset} & Training Set & Test Set \\
\midrule
\midrule
\multirow{7}{*}{\makecell{NASA\\(OC = 20)}} & \multirow{4}{*}{SOH} & B0005 B0006 B0007 & B0018 \\
                             &     & B0005 B0006 B0018 & B0007 \\
                             &     & B0005 B0007 B0018 & B0006 \\
                             &     & B0006 B0007 B0018 & B0005 \\
\cmidrule(lr){2-4}
                             & \multirow{3}{*}{RUL} & B0005 B0006       & B0018 \\
                             &     & B0005 B0018       & B0006 \\
                             &     & B0006 B0018       & B0005 \\
\midrule
\midrule
\multirow{4}{*}{\makecell{CALCE\\(OC = 50)}} & \multirow{4}{*}{SOH / RUL} & CS2\_35 CS2\_36 CS2\_37 & CS2\_38 \\
                              &         & CS2\_35 CS2\_36 CS2\_38 & CS2\_37 \\
                              &         & CS2\_35 CS2\_37 CS2\_38 & CS2\_36 \\
                              &         & CS2\_36 CS2\_37 CS2\_38 & CS2\_35 \\
\midrule
\midrule
\multirow{4}{*}{\makecell{XJTU\\(OC=20)}} & \multirow{4}{*}{SOH / RUL} & Batch1\_1 Batch1\_2 Batch1\_3 & Batch1\_4 \\
                             &         & Batch1\_1 Batch1\_2 Batch1\_4 & Batch1\_3 \\
                             &         & Batch1\_1 Batch1\_3 Batch1\_4 & Batch1\_2 \\
                             &         & Batch1\_2 Batch1\_3 Batch1\_4 & Batch1\_1 \\
\midrule
\midrule
\multirow{4}{*}{\makecell{Oxford\\(OC = 10)}} & \multirow{4}{*}{SOH / RUL} & Ox\_1 Ox\_3 Ox\_6 & Ox\_7 \\
                               &         & Ox\_1 Ox\_3 Ox\_7 & Ox\_6 \\
                               &         & Ox\_1 Ox\_6 Ox\_7 & Ox\_3 \\
                               &         & Ox\_3 Ox\_6 Ox\_7 & Ox\_1 \\
\midrule
\midrule
\multirow{8}{*}{\makecell{MIT\\(OC = 40)}} & \multirow{4}{*}{SOH} & 10\_CH44 30\_CH30 50\_CH32 & 70\_CH46 \\
                            &     & 10\_CH44 30\_CH30 70\_CH46 & 50\_CH32 \\
                            &     & 10\_CH44 50\_CH32 70\_CH46 & 30\_CH30 \\
                            &     & 30\_CH30 50\_CH32 70\_CH46 & 10\_CH44 \\
\cmidrule(lr){2-4}
                            & \multirow{3}{*}{RUL} & 10\_CH44 30\_CH30 70\_CH46 & 70\_CH46 \\
                            &     & 10\_CH44 70\_CH46 30\_CH30 & 30\_CH30 \\
                            &     & 30\_CH30 70\_CH46 10\_CH44 & 10\_CH44 \\
\bottomrule[0.5mm]
\label{Data_partitioning}
\end{tabular}
\end{table}

\noindent\textbf{Step B:} Constructing the joint SOH and RUL prediction NN. The network architecture includes a feature extraction module, a temporal modeling module, a dual-stream attention module, and task layer connected in sequence, as outlined in Section \ref{method}. Once the network is constructed, the Hyperopt optimizer is employed to automatically search for optimal hyperparameters. This optimizer demonstrates superior efficiency compared to the Bayesian optimizer based on Gaussian processes. In this network, the number of neurons for the FFN is determined to be 64, the number of neurons in the task layer is 128, the number of sparse attention heads is 4, and the number of neurons in the extended LSTM is 128.

\noindent\textbf{Step C:} Training the proposed network. The training data are fed into the network, which is trained using the Adam optimizer, with the mean square error (MSE) as the loss function. The Hyperopt method is also used to optimize the training hyperparameters, including the number of epochs, learning rate, and batch size. To balance efficiency and accuracy, the number of epochs is set to 50, and the batch size is set to 32. To accelerate model convergence and prevent overfitting, the initial learning rate is set to $\left( {1{\rm{e}} - 4} \right) * \frac{1}{8}$ and warmed up to $\left( {1e - 4} \right)$ after 7 epochs. The learning rate is then reduced to 75\% of the previous value at each subsequent epoch. Once training is complete, the model is saved for further use.

\noindent\textbf{Step D:} The test set is fed into the well-trained network for evaluation. To quantitatively assess the network's performance, various evaluation metrics are employed and calculated as follows:

\begin{equation}
\left\{ {\begin{array}{*{20}{c}}
{\mathrm{MAE} = \frac{1}{n}\sum\limits_{i = 1}^n {\left| {{{\hat y}_i} - {y_i}} \right|} }\\
{\mathrm{RMSE} = \sqrt {\frac{1}{n}\sum\limits_{i = 1}^n {{{\left( {{{\hat y}_i} - {y_i}} \right)}^2}} } }\\
{\mathrm{MAPE} = \frac{1}{n}\sum\limits_{i = 1}^n {\left| {\frac{{{{\hat y}_i} - {y_i}}}{{{y_i}}}} \right|} }\\
{\mathrm{MedAE} = median\left( {\left| {{{\hat y}_i} - {y_i}} \right|,...,\left| {{{\hat y}_n} - {y_n}} \right|} \right)}
\end{array}} \right.
\end{equation}
where $n$ is the number of cycles. For SOH prediction, Mean Absolute Error (MAE), Root Mean Square Error (RMSE) and Mean Absolute Percentage Error (MAPE) are used, and ${\hat y_i}$ and ${y_i}$ represent the real and predicted capacity of the battery. For RUL prediction, MAE, RMSE and Median Absolute Error (MedAE) are used, and ${\hat y_i}$ and ${y_i}$ represent the real and predicted RUL of the battery.

\section{Experimental results and analysis}
\label{5}

\subsection{Experimental platform}
\label{6}
The experimental setup is conducted on a desktop system with the following specifications: CPU: Intel(R) Core(TM) i9-14900HX 2.20 GHz, Memory: 32 GB, and GPU: NVIDIA GeForce RTX 4060. The models used in the experiment are implemented using Python 3.8 and developed on the PyTorch framework.

\begin{table*}[t]
    \centering
    \caption{The ablation experiment results for the proposed method in this paper, illustrate the average MAE and RMSE across four cells for both datasets.}
    \begin{tabular}{c cccc cccc}
     \toprule[0.5mm]
     & \multicolumn{3}{c}{Network} & & \multicolumn{4}{c}{Evaluation Criteria}\\
     \cmidrule{2-4} \cmidrule{6-9} 
     & FEM & IE-LSTM & DSAM & & MAE (\%, SOH) & RMSE (\%, SOH) & MAE (RUL) & RMSE (RUL) \\
     \midrule
     \midrule
      \multirow{4}{*}{\rotatebox{0}{NASA dataset}}
     & \checkmark & \checkmark &   & & 0.0082 & 0.0111 & 5.04 & 6.82 \\
     & \checkmark &  &  \checkmark & & 0.0145 & 0.0180 & 4.97 & 6.31 \\
     &  & \checkmark &  \checkmark & & 0.0112 & 0.0145 & 4.53 & 6.71 \\
     & \checkmark & \checkmark &  \checkmark & & 0.0053 & 0.0084 & 4.01 & 5.39 \\
     \midrule
     \midrule
     \multirow{4}{*}{\rotatebox{0}{CALCE dataset}}
     & \checkmark & \checkmark &   & & 0.0099 & 0.0126 & 46.39 & 61.87 \\
     & \checkmark &  &  \checkmark & & 0.0103 & 0.0130 & 51.07 & 66.78 \\
     &  & \checkmark &  \checkmark & & 0.0122 & 0.0160 & 47.79 & 63.62 \\
     & \checkmark & \checkmark &  \checkmark & & 0.0068 & 0.0093 & 40.27 & 53.35 \\
     \midrule
     \midrule
     \multirow{4}{*}{\rotatebox{0}{XJTU dataset}}
     & \checkmark & \checkmark &   & & 0.99 & 1.25 & 47.96 & 64.12 \\
     & \checkmark &  &  \checkmark & & 0.77 & 0.96 & 28.81 & 38.49 \\
     &  & \checkmark & \checkmark & & 0.64 & 0.80 & 12.57 & 15.77 \\
     & \checkmark & \checkmark &  \checkmark & & 0.47 & 0.62 & 16.94 & 21.05 \\
     \midrule
     \midrule
     \multirow{4}{*}{\rotatebox{0}{Oxford dataset}}
     & \checkmark & \checkmark &   & & 1.34 & 1.55 & 1.52 & 2.36 \\
     & \checkmark &  &  \checkmark & & 0.93 & 1.07 & 0.80 & 1.17 \\
     &  & \checkmark & \checkmark & & 0.84 & 0.93 & 0.83 & 1.19 \\
     & \checkmark & \checkmark &  \checkmark & & 0.38 & 0.46 & 0.68 & 1.05 \\
     \midrule
     \midrule
     \multirow{4}{*}{\rotatebox{0}{MIT dataset}}
     & \checkmark & \checkmark &   & & 2.26 & 2.76 & 30.70 & 34.91 \\
     & \checkmark &  & \checkmark & & 1.24 & 1.59 & 23.60 & 28.95 \\
     &  & \checkmark & \checkmark & & 2.47 & 3.69 & 31.42 & 37.63 \\
     & \checkmark & \checkmark &  \checkmark & & 0.78 & 1.09 & 19.22 & 24.94 \\

    \bottomrule[0.5mm]
    \end{tabular}
    \label{tableab}
\end{table*}

\begin{figure*}[t]
    \centering
    \includegraphics[width=1\linewidth]{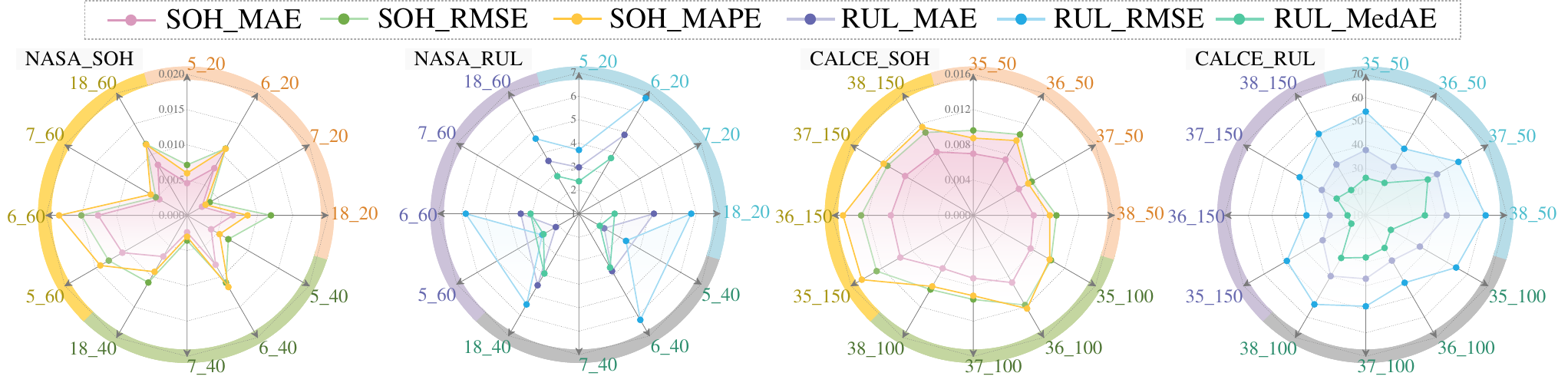}
    \caption{Experimental results under different observation cycles on the NASA and CALCE datasets.}
    \label{figuresp}
    \vspace{-7.5pt}
\end{figure*}

\begin{figure*}[t]
    \centering
    \includegraphics[width=0.9\linewidth]{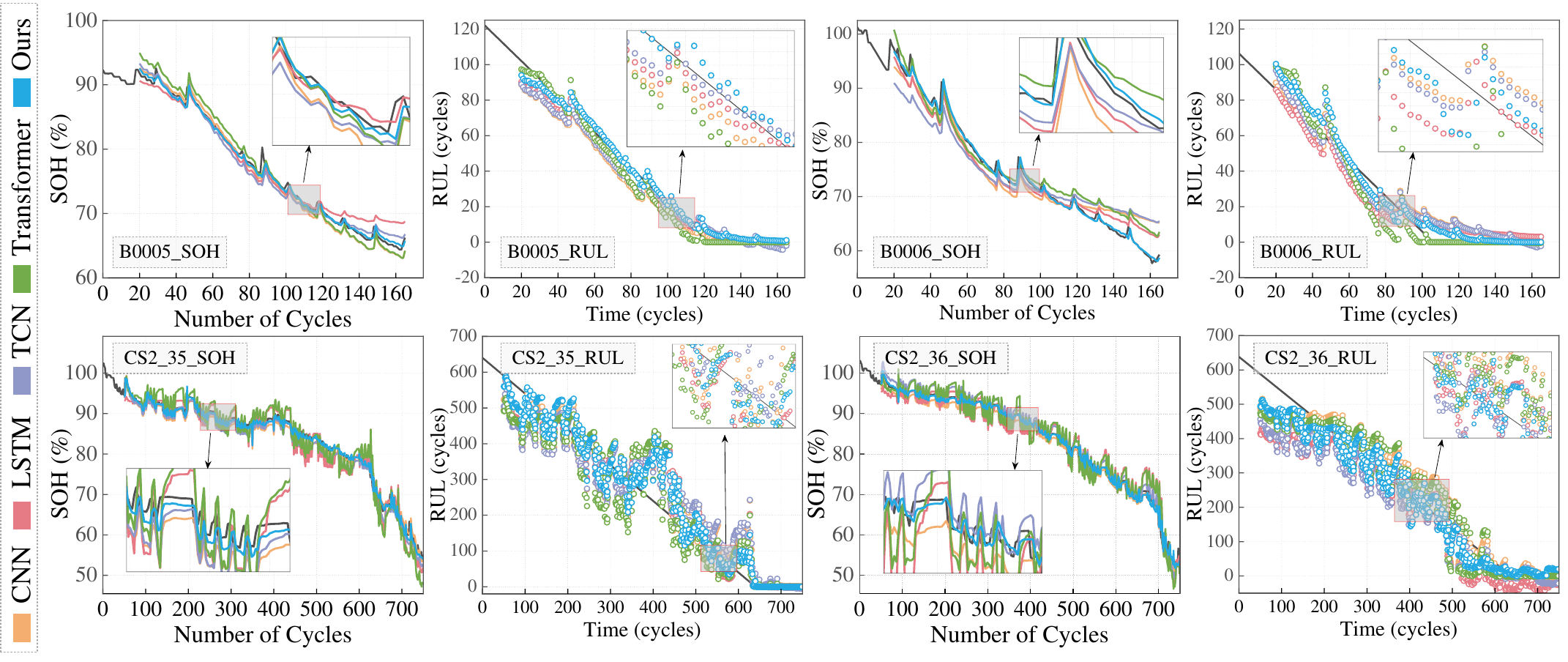}
    \caption{Comparative results of the proposed method versus other NN-based methods.}
    \label{figurecompare}
    \vspace{-7.5pt}
\end{figure*}

\begin{table*}[t]
    \centering
    \caption{Comparison results are presented between our method and both generic and state-of-the-art methods. While CNN, LSTM, TCN, and Transformer models are evaluated on our platform, other comparative values are directly sourced from the experimental results reported in relevant papers. The three evaluation metrics for SOH are displayed as percentages, where lower values indicate improved predictive performance. The best results are highlighted in \textbf{bold}.}
    \resizebox{2\columnwidth}{!}{
    \begin{tabular}{cc cccc cccc cccc ccc}
     \toprule[0.5mm]
     \multirow{1}{*}{\rotatebox{0}{\textbf{SOH}}}  & &  \multicolumn{3}{c}{B0005} & & \multicolumn{3}{c}{B0006} & &  \multicolumn{3}{c}{CS2\_35} & & \multicolumn{3}{c}{CS2\_36}\\
     \cmidrule{3-5} \cmidrule{7-9} \cmidrule{11-13} \cmidrule{15-17}
     Method & & MAE  & RMSE  & MAPE & & MAE  & RMSE  & MAPE  & & MAE  & RMSE  & MAPE & & MAE  & RMSE  & MAPE\\
     \midrule
     \multirow{1}{*}{\rotatebox{0}{CNN}}
     & & 0.87 & 1.08 & 1.18 & & 2.29 & 2.93 & 3.31 &  & 1.34 & 1.59 & 1.62 &  & 1.33 & 1.52 & 1.63 \\
     \multirow{1}{*}{\rotatebox{0}{LSTM}}
     & & 1.21 & 1.66 & 1.69 & & 1.84 & 2.24 & 2.56 &  & 2.05 & 2.57 & 2.59 &  & 2.05 & 2.64 & 2.61 \\
     \multirow{1}{*}{\rotatebox{0}{TCN}}
     & & 0.95 & 1.20 & 1.26 & & 3.09 & 3.70 & 4.24 &  & 1.43 & 1.74 & 1.78 &  & 1.56 & 1.86 & 1.93 \\
     \multirow{1}{*}{\rotatebox{0}{Transformer}}
     & & 1.20 & 1.42 & 1.53 & & 2.06 & 2.52 & 2.99 &  & 2.24 & 2.83 & 2.87 &  & 2.15 & 2.68 & 2.73 \\
     
     \multirow{1}{*}{\rotatebox{0}{\cite{c3}}}
     & & 0.89 & 1.43 & \textbf{0.57} & & / & / & / &  & 0.87 & 1.57 & 1.42 &  & 0.90 & 1.42 & 1.45 \\
     \multirow{1}{*}{\rotatebox{0}{\cite{c4}}}
     & & / & / & / & & / & / & / &  & 1.00 & 1.60 & 1.70 &  & 2.00 & 2.40 & 4.00 \\
     \multirow{1}{*}{\rotatebox{0}{\cite{c_add_1}}}
     & & 1.29 & 1.70 & / & & 1.74 & 2.48 & / &  & / & / & / &  & / & / & / \\
     \multirow{1}{*}{\rotatebox{0}{\cite{c_add_2}}}
     & & 0.50 & 0.80 & 0.83 & & 1.50 & 1.90 & 1.98 &  & / & / & / &  & / & / & / \\
     \multirow{1}{*}{\rotatebox{0}{\cite{c_add_4}}}
     & & 0.76 & 1.40 & 0.87 & & 0.82 & 1.35 & 0.87 &  & 0.91 & 1.40 & 0.87 &  & 0.87 & 1.47 & 0.98 \\

     \multirow{1}{*}{\rotatebox{0}{Ours}}
     & & \textbf{0.46} & \textbf{0.72} & 0.60 & & \textbf{0.77} & \textbf{1.09} & \textbf{1.09} &  & \textbf{0.70} & \textbf{0.96} & \textbf{0.88} &  & \textbf{0.73} & \textbf{1.06} & \textbf{0.98} \\
     
     \midrule
     \midrule
     
     \multirow{1}{*}{\rotatebox{0}{\textbf{RUL}}}  & &  \multicolumn{3}{c}{B0005} & & \multicolumn{3}{c}{B0006} & &  \multicolumn{3}{c}{CS2\_35} & & \multicolumn{3}{c}{CS2\_36}\\
     \cmidrule{3-5} \cmidrule{7-9} \cmidrule{11-13} \cmidrule{15-17}
     Method & & MAE  & RMSE  & MedAE & & MAE  & RMSE  & MedAE  & & MAE  & RMSE  & MedAE & & MAE  & RMSE  & MedAE\\
     
     \midrule
     \multirow{1}{*}{\rotatebox{0}{CNN}}
     &  & 5.29 & 6.96 & 2.38 &  & 5.73 & 6.78 & 3.73 & & 52.31 & 69.61 & 44.64 & & 41.43 & 52.51 & 33.17  \\
     \multirow{1}{*}{\rotatebox{0}{LSTM}}
     &  & 4.42 & 5.50 & 3.60 &  & 6.49 & 7.55 & 5.71 & & 45.85 & 63.84 & 34.87 & & 56.50 & 72.30 & 40.97 \\
     \multirow{1}{*}{\rotatebox{0}{TCN}}
     &  & 5.30 & 6.50 & 4.24 &  & 5.89 & 6.86 & 5.77 & & 54.42 & 71.61 & 45.02 & & 48.10 & 64.43 & 32.87 \\
     \multirow{1}{*}{\rotatebox{0}{Transformer}}
     &  & 3.14 & 4.60 & \textbf{2.04} &  & 5.86 & 9.24 & \textbf{1.72} & & 53.81 & 73.17 & 41.23 & & 42.76 & 55.40 & 36.43  \\
     \multirow{1}{*}{\rotatebox{0}{\cite{c5}}}
     & & 7.83 & 9.97 & / & & / & / & / &  & / & / & / &  & / & / & / \\
     \multirow{1}{*}{\rotatebox{0}{\cite{c6}}}
     & & 11.89 & 12.31 & 12.38 & & 14.01 & 16.94 & 11.13 &  & / & / & / &  & / & / & / \\
     \multirow{1}{*}{\rotatebox{0}{Ours}}
     &  & \textbf{2.97} & \textbf{3.71} & 2.38 &  & \textbf{4.87} & \textbf{6.67} & 3.73 & & \textbf{37.78} & \textbf{54.15} & \textbf{25.96} & & \textbf{33.86} & \textbf{42.72} & \textbf{25.96} \\
     \bottomrule[0.5mm]
     \end{tabular}}
     \label{tablepaper}
\end{table*}

\begin{figure*}[t]
    \centering
    \includegraphics[width=0.9\linewidth]{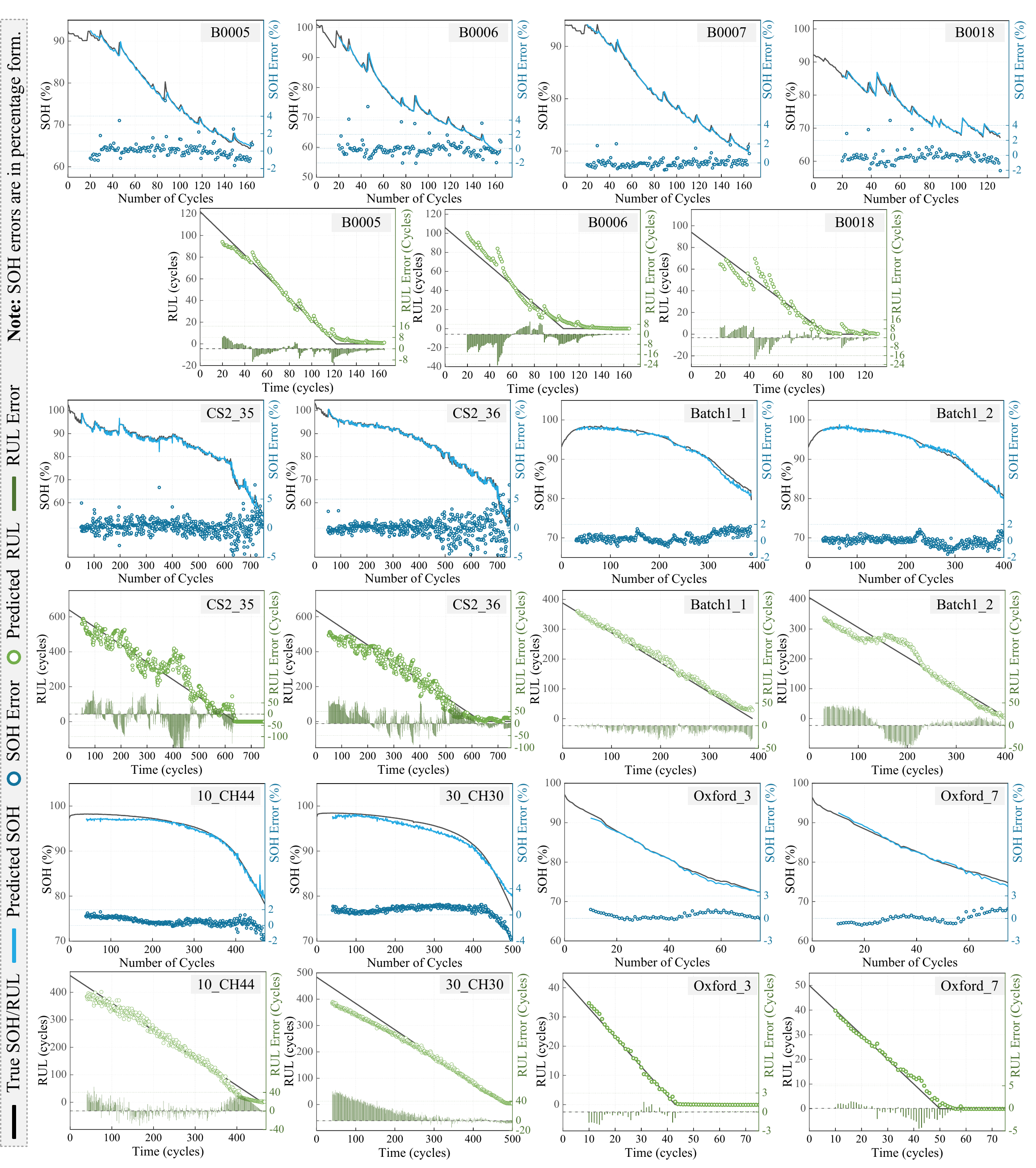}
    \caption{The SOH and RUL prediction results of the proposed method for five datasets are presented.}
    \label{figureall}
    \vspace{-7.5pt}
\end{figure*}

\subsection{Ablation experiments}
To assess the effectiveness and generalizability of each module, ablation experiments are conducted on all five datasets. The quantitative results are summarized in Table \ref{tableab}, where values represent the average performance across all batteries. The results demonstrate that, across all datasets, the full method incorporating all modules consistently achieves the best performance. In contrast, the removal of the IE-LSTM or DSAM modules leads to a substantial decline in prediction accuracy. The IE-LSTM module is essential for capturing temporal dependencies, and the absence of this module would compromise the preservation of critical information. Similarly, the DSAM module adaptively assigns weights to the two tasks, enhancing feature sensitivity and supporting targeted learning. Its absence results in the neglect of critical features, producing overly “averaged” predictions. Specifically, for the NASA dataset, excluding the FEM, IE-LSTM, and DSAM modules increase the average SOH prediction MAE by 111.0\%, 164.9\%, and 53.7\%, respectively, and the average RUL prediction RMSE by 24.6\%, 17.1\%, and 26.6\%, respectively. On the CALCE data set, the MAEs of SOH increase by 71. 5\%, 39. 3\%, and 35. 1\%, while the RMSEs of RUL increase by 19. 2\%, 25. 2\%, and 16. 0\%, respectively. For the XJTU dataset, the corresponding increases are 36. 2\%, 63. 8\% and 110. 6\% for SOH, and 25. 1\%, 82. 9\% and 204. 6\% for RUL. On the Oxford dataset, the average increases are 121.4\%, 146.4\%, and 253.6\% for SOH, and 22.1\%, 17.6\%, and 123.5\% for RUL. Finally, in the MIT dataset, the MAEs of SOH increased by 216. 7\%, 59. 0\% and 187. 7\%, while the RMSEs of RUL increased by 50. 9\%, 16. 1\%, and 40. 0\%, respectively. These findings further validate the effectiveness of the three proposed modules.

\subsection{Experiments on observed cycles}
\label{7}
Fig. \ref{figuresp} displays the quantitative SOH and RUL prediction metrics for experiments with various observation cycles across eight batteries: NASA B0005, B0006, B0007, B0018 and CALCE CS2\_35, CS2\_36, CS2\_37, CS2\_38. In the polar plots, the notation “a\_b” indicates the cell and the observation cycle, where “a” refers to the specific cell and “b” denotes the observation cycle (e.g., “5\_20” represents the prediction result of battery B0005 at the 20-th cycle). Since battery B0007 does not reach its EOL, it is excluded from the RUL prediction; hence, its results are left blank in the NASA\_RUL plot. Different colors correspond to different evaluation metrics. The proposed method consistently achieves superior results, independent of the starting point of the observation cycles. Specifically, the average MAE, RMSE and MAPE for SOH prediction are 0.65\%, 0.92\% and 0.90\% for the NASA cells, and 0.78\%, 1.05\% and 1.07\% for the CALCE cells. For RUL prediction, the average MAE, RMSE and MedAE are 3.54, 4.94 and 2.99 on the NASA cells, and 35.81, 49.15, and 26.36 on the CALCE cells. These results demonstrate the method’s capacity for accurate early-stage battery state prediction. However, it is important to note that the RUL predictions may exhibit mid-term bias, as the dataset includes extended resting periods before EOL, amplifying the capacity regeneration phenomenon and introducing greater error. This effect similarly impacts SOH prediction, resulting in slightly elevated errors. Nevertheless, the proposed approach delivers consistently improved predictions across all observation cycle durations.

\subsection{Comparison among different NN methods}

\subsubsection{Comparison with conventional methods}
\label{NNmethod}

Comparison experiments are conducted on the NASA B0005, B0006 and CALCE CS2\_35, CS2\_36 batteries using several networks: conventional algorithms (CNN and LSTM) and state-of-the-art networks (TCN and Transformer). The results are presented in Fig. \ref{figurecompare}, with quantitative metrics displayed in Table \ref{tablepaper}. For SOH prediction, all NN-based methods capture the battery degradation trend, but the proposed method significantly enhances prediction accuracy, achieving average improvements of 161.0\%, 123.4\%, and 158.7\% in MAE, RMSE, and MAPE, respectively, over the other networks. The NASA dataset, particularly B0018, exhibits pronounced capacity regeneration phenomena, which impose higher demands on the network’s capability to capture corresponding relationships. The proposed method leverages its improved exponential gating mechanism and novel data storage strategy to adaptively refine storage decisions based on degradation information, thereby enhancing both modeling capacity and adaptability. For the CALCE dataset, the raw voltage curves show considerable fluctuations and significant discrepancies across different cycles. By incorporating the DSAM module, the proposed method assigns higher weights to critical feature segments while ignoring non-essential ones, allowing it to effectively capture distinct key features across cycles.

For RUL prediction, the proposed method yields average improvements of 36.1\%, 32.4\%, and 34.2\% in MAE, RMSE, and MedAE, respectively, compared to other networks. The limitations of existing models become apparent in this context: CNN struggles with time-series modeling, while LSTM suffers from inherent memory constraints, both leading to reduced performance on multi-series prediction tasks. TCN, although effective in capturing temporal patterns, lacks dynamic adaptability and fails to model instantaneous changes flexibly. Transformer-based methods generally achieve strong results across diverse domains, yet their reliance on large training datasets reduces effectiveness in battery prognostics, where data volume is inherently limited. Overall, conventional networks often encounter information interference when addressing multi-time-series prediction tasks due to the absence of targeted learning mechanisms. By contrast, the DSAM module in the proposed network achieves superior task-specific performance by leveraging an independent attention mechanism to selectively capture critical features.

\subsubsection{Comparison with other state-of-the-art methods in the literature}

To demonstrate the advancement of the proposed method, a comparative analysis is conducted against recently published studies. To ensure consistency and avoid the confounding effects of differing experimental platforms and configuration parameters, the results from the published studies are directly utilized for comparison. These results, shown in Table \ref{tablepaper}, indicate that the proposed method achieves the highest SOH and RUL prediction accuracy in most cases, further validating its superiority.

\begin{figure*}[t]
    \centering
    \includegraphics[width=1\linewidth]{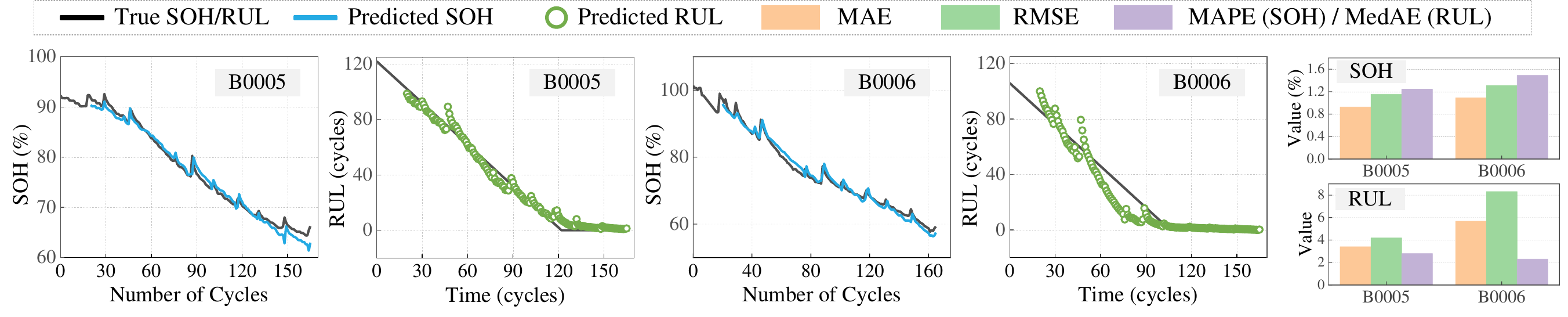}
    \caption{Experimental results with non-full data as input to the proposed network.}
    \label{factor}
    \vspace{-7.5pt}
\end{figure*}

\begin{table*}[t]
    \centering
    \caption{Quantitative results of the method proposed in this paper on all eight cells, where the evaluation metrics for SOH are presented in percentage form, and  M/M represents MAPE at SOH task, MedAE at RUL task.}
    \begin{tabular}{cc cccc cccc cccc ccc}
     \toprule[0.5mm]
     \multirow{1}{*}{\rotatebox{0}{\textbf{Task}}}  & &  \multicolumn{3}{c}{B0005} & & \multicolumn{3}{c}{B0006} & &  \multicolumn{3}{c}{B0007} & & \multicolumn{3}{c}{B00018}\\
     \cmidrule{3-5} \cmidrule{7-9} \cmidrule{11-13} \cmidrule{15-17}
     Metrics & & MAE  & RMSE  & M/M & & MAE  & RMSE  & M/M  & & MAE  & RMSE  & M/M & & MAE  & RMSE  & M/M\\
     \midrule
     \multirow{1}{*}{\rotatebox{0}{SOH (\%)}}
     & & 0.46 & 0.72 & 0.60 & & 0.77 & 1.09 & 1.09 &  & 0.24 & 0.37 & 0.30 &  & 0.65 & 1.19 & 0.86 \\
     \multirow{1}{*}{\rotatebox{0}{RUL}}
     & & 2.97 & 3.71 & 2.38 & & 4.87 & 6.67 & 3.73 &  & / & / & / &  & 4.19 & 5.78 & 2.52 \\
     
     \midrule
     \midrule
     & & \multicolumn{3}{c}{CS2\_35} & & \multicolumn{3}{c}{CS2\_36} & &  \multicolumn{3}{c}{CS2\_37} & & \multicolumn{3}{c}{CS2\_38}\\
     \cmidrule{3-5} \cmidrule{7-9} \cmidrule{11-13} \cmidrule{15-17}
     \multirow{1}{*}{\rotatebox{0}{SOH (\%)}}
     & & 0.70 & 0.96 & 0.88 & & 0.73 & 1.06 & 0.98 &  & 0.60 & 0.76 & 0.72 &  & 0.69 & 0.94 & 0.87 \\
     \multirow{1}{*}{\rotatebox{0}{RUL}}
     & & 37.78 & 54.15 & 25.96 & & 33.86 & 42.72 & 25.96 &  & 45.01 & 55.53 & 40.46 &  & 44.42 & 60.99 & 35.17 \\

     \midrule
     \midrule
     & & \multicolumn{3}{c}{Batch1\_1} & & \multicolumn{3}{c}{Batch1\_2} & &  \multicolumn{3}{c}{Batch1\_3} & & \multicolumn{3}{c}{Batch1\_4}\\
     \cmidrule{3-5} \cmidrule{7-9} \cmidrule{11-13} \cmidrule{15-17}
     \multirow{1}{*}{\rotatebox{0}{SOH (\%)}}
     & & 0.45 & 0.59 & 0.50 & & 0.36 & 0.48 & 0.39 &  & 0.45 & 0.62 & 0.50 &  & 0.62 & 0.76 & 0.67 \\
     \multirow{1}{*}{\rotatebox{0}{RUL}}
     & & 11.20 & 13.30 & 10.82 & & 21.89 & 27.01 & 16.76 &  & 14.89 & 20.39 & 11.59 &  & 19.73 & 23.49 & 18.47 \\

     \midrule
     \midrule
     & & \multicolumn{3}{c}{Oxford\_1} & & \multicolumn{3}{c}{red}{Oxford\_3} & &  \multicolumn{3}{c}{Oxford\_6} & & \multicolumn{3}{c}{Oxford\_7}\\
     \cmidrule{3-5} \cmidrule{7-9} \cmidrule{11-13} \cmidrule{15-17}
     \multirow{1}{*}{\rotatebox{0}{SOH (\%)}}
     & & 0.35 & 0.41 & 0.58 & & 0.35 & 0.45 & 0.59 &  & 0.39 & 0.45 & 0.63 &  & 0.44 & 0.51 & 0.72 \\
     \multirow{1}{*}{\rotatebox{0}{RUL}}
     & & 0.46 & 0.75 & 0.13 & & 0.44 & 0.71 & 0.10 &  & 0.83 & 1.29 & 0.42 &  & 1.00 & 1.44 & 0.82 \\

     \midrule
     \midrule
     & & \multicolumn{3}{c}{10\_CH44} & & \multicolumn{3}{c}{30\_CH30} & &  \multicolumn{3}{c}{50\_CH32} & & \multicolumn{3}{c}{70\_CH46}\\
     \cmidrule{3-5} \cmidrule{7-9} \cmidrule{11-13} \cmidrule{15-17}
     \multirow{1}{*}{\rotatebox{0}{SOH (\%)}}
     & & 0.69 & 0.97 & 0.76 & & 0.95 & 1.12 & 1.04 &  & 0.70 & 1.16 & 0.77 &  & 0.80 & 1.09 & 0.88 \\
     \multirow{1}{*}{\rotatebox{0}{RUL}}
     & & 13.46 & 17.34 & 10.67 & & 18.64 & 26.93 & 10.55 &  & / & / & / &  & 25.56 & 30.55 & 23.71 \\
    
     \bottomrule[0.5mm]
     \end{tabular}
     \label{tableall}
\end{table*}

\subsection{SOH and RUL prediction}
The proposed method is evaluated on five battery datasets, with four cells selected from each dataset, resulting in a total of 20 cells. The experimental results are presented in Figure \ref{figureall}, while additional details are provided in Supplementary Material \uppercase\expandafter{\romannumeral2}. The error of SOH is calculated as $\mathrm{SO}{\mathrm{H}_{Error}} = \left( {\frac{{{C_{real}} - {C_{pre}}}}{{{C_{real}}}}} \right) \times 100\% $ and the error of RUL is calculated as $\mathrm{RU{L}_{Error}} = \mathrm{RU}{\mathrm{L}_{real}} - \mathrm{RU}{\mathrm{L}}_{pre}$. The visualized prediction results clearly demonstrate that the proposed method can accurately predict SOH and RUL across different datasets. For the NASA and Oxford datasets, which contain dense voltage data points and long charge–discharge durations within each cycle, the IE-LSTM effectively captures long-term dependencies through temporal modeling. For the XJTU dataset, where the raw voltage data points are relatively sparse, the proposed FEM extracts essential features more effectively, facilitating subsequent network-based relational modeling. For the CALCE and MIT datasets, where the plateau phases in voltage curves are less pronounced and key feature information is more dispersed, the proposed DSAM leverages dual-stream attention to assign higher weights to critical segments in the raw data, enabling precise capture of crucial information.

The quantitative evaluation results are summarized in Table \ref{tableall}. For SOH estimation, the proposed method achieves average MAE values of 0.53\%, 0.68\%, 0.47\%, 0.38\%, and 0.78\% on the NASA, CALCE, XJTU, Oxford, and MIT datasets, respectively. For RUL estimation, the corresponding average MAE values are 4.01, 40.27, 16.94, 0.68, and 19.22, respectively. It is important to note that RUL prediction errors are typically expressed as absolute values. Due to the significant variation in the EOL cycle counts among different datasets, the range of error intervals also varies accordingly.

\subsection{Non-full charge data experiment}
\label{Non-full}
In practical battery usage scenarios, obtaining complete discharge cycles is often difficult. To account for this and to further validate the practical applicability of the proposed method, this section uses the B0005 and B0006 batteries as case studies. Feature factors are extracted directly from the original charge–discharge curves and employed as network inputs for subsequent prediction tasks. Feature selection is guided by the following principles: (1) consistent with previous studies \cite{R21}, features from the low-voltage stage are utilized; (2) to address inconsistent sequence lengths, feature factors are used to substitute portions of the charge–discharge data; (3) Pearson correlation analysis is applied to identify and retain feature factors that exhibit strong correlations with capacity degradation. Specifically, four key voltage-related features were identified: (1) constant current charging voltage onset to peak time, (2) the duration of the 3.9V–4.1V voltage plateau, (3) the slope of the voltage rise within the 3.6V–4.0V range, and (4) the voltage-time integral during the CC phase.

The experimental results, presented in Fig. \ref{factor}, show that the model can still achieve accurate predictions of both SOH and RUL when using these extracted features instead of full-cycle voltage data. For SOH estimation, the model achieves average MAE, RMSE, and MAPE values of 1.00\%, 1.23\%, and 1.37\%, respectively. For RUL estimation, the corresponding MAE, RMSE, and MedAE values are 4.54, 6.25, and 2.55. These results confirm that, given adequate training data, the proposed network is capable of jointly predicting multiple time-dependent battery performance indicators with high accuracy.

\section{Conclusion}
\label{5}
This paper proposes a joint prediction approach for battery SOH and RUL, integrating the multi-scale FEM, the IE-LSTM, and the DSAM. The method begins with a FEM to capture original feature details and deeply extract battery degradation patterns. Next, the LSTM network is improved to strengthen memory capabilities, optimizing it for long-term battery degradation modeling and effective temporal information processing. DSAM is then constructed to selectively extract critical information specific to the SOH and RUL tasks, enabling targeted multi-task learning. Finally, a dual-task layer performs a multi-to-two mapping to jointly predict SOH and RUL. Network and training hyperparameters are tuned using an automated optimization search to minimize manual setup complexity. A comprehensive series of experiments was conducted using multiple battery aging datasets. The results demonstrated that the proposed method achieved average MAE, RMSE, and MAPE/MedAE values below 0.60\%, 0.76\%, and 0.72\%, respectively, for SOH estimation, and below 16.74, 21.83, and 13.37 for RUL prediction. This paper also observed that prediction errors tend to increase as the battery approaches EOL, likely due to the extended resting times of batteries nearing EOL. However, the accuracy of RUL predictions significantly improves once the battery reaches EOL, as the RULs are constant (i.e., zero). This constancy simplifies the learning of mapping relationships, enhancing the network’s predictive performance at EOL.

\noindent\textbf{More Discussion.} In future work, our research will focus on enhancing the practical applicability of the proposed method, with particular attention to three key areas. First, given that real-world applications often involve large-capacity batteries, it is crucial to acquire long-term cyclic charge/discharge data specific to such batteries. Second, the practical value of the method can be further validated through the analysis of extensive EV operational data spanning more than five years. Third, since capacity degradation trends vary under conditions such as battery decommissioning or resale, future efforts will incorporate the prediction of battery life during the later stages of use and under diverse operating conditions.

\small
\bibliographystyle{IEEEtran}
\bibliography{IEEEabrv, sample}

\begin{thebibliography}{10}
\providecommand{\url}[1]{#1}
\csname url@samestyle\endcsname
\providecommand{\newblock}{\relax}
\providecommand{\bibinfo}[2]{#2}
\providecommand{\BIBentrySTDinterwordspacing}{\spaceskip=0pt\relax}
\providecommand{\BIBentryALTinterwordstretchfactor}{4}
\providecommand{\BIBentryALTinterwordspacing}{\spaceskip=\fontdimen2\font plus
\BIBentryALTinterwordstretchfactor\fontdimen3\font minus \fontdimen4\font\relax}
\providecommand{\BIBforeignlanguage}[2]{{%
\expandafter\ifx\csname l@#1\endcsname\relax
\typeout{** WARNING: IEEEtran.bst: No hyphenation pattern has been}%
\typeout{** loaded for the language `#1'. Using the pattern for}%
\typeout{** the default language instead.}%
\else
\language=\csname l@#1\endcsname
\fi
#2}}
\providecommand{\BIBdecl}{\relax}
\BIBdecl

\bibitem{r1_3}
C.~Zhang, S.~Zhao, Z.~Yang, and Y.~He, ``A multi-fault diagnosis method for lithium-ion battery pack using curvilinear manhattan distance evaluation and voltage difference analysis,'' \emph{Journal of Energy Storage}, vol.~67, p. 107575, 2023.

\bibitem{1}
Q.~Yu, C.~Wang, J.~Li, R.~Xiong, and M.~Pecht, ``Challenges and outlook for lithium-ion battery fault diagnosis methods from the laboratory to real world applications,'' \emph{ETransportation}, vol.~17, p. 100254, 2023.

\bibitem{2}
L.~Shen, J.~Li, L.~Meng, L.~Zhu, and H.~T. Shen, ``Transfer learning-based state of charge and state of health estimation for li-ion batteries: A review,'' \emph{IEEE Transactions on Transportation Electrification}, vol.~10, no.~1, pp. 1465--1481, 2023.

\bibitem{4}
C.~Wang, Z.~Bao, H.~Lin, Z.~He, and M.~Gao, ``An exponential transformer for learning interpretable temporal information in remaining useful life prediction of lithium-ion battery,'' \emph{IEEE Transactions on Transportation Electrification}, 2025.

\bibitem{7}
L.~Chen, S.~Xie, A.~M. Lopes, H.~Li, X.~Bao, C.~Zhang, and P.~Li, ``A new soh estimation method for lithium-ion batteries based on model-data-fusion,'' \emph{Energy}, vol. 286, p. 129597, 2024.

\bibitem{8}
W.~Hu and Q.~Qian, ``Lithium-ion battery state of health and failure analysis with mixture weibull and equivalent circuit model,'' \emph{iScience}, vol.~27, no.~6, 2024.

\bibitem{9}
S.~Chen, Q.~Zhang, F.~Wang, D.~Wang, and Z.~He, ``An electrochemical-thermal-aging effects coupled model for lithium-ion batteries performance simulation and state of health estimation,'' \emph{Applied Thermal Engineering}, vol. 239, p. 122128, 2024.

\bibitem{10}
G.~Zhu, C.~Kong, J.~V. Wang, J.~Kang, Q.~Wang, and C.~Qian, ``A fractional-order electrochemical lithium-ion batteries model considering electrolyte polarization and aging mechanism for state of health estimation,'' \emph{Journal of Energy Storage}, vol.~72, p. 108649, 2023.

\bibitem{r3_4}
V.~Sulzer, P.~Mohtat, S.~Pannala, J.~B. Siegel, and A.~G. Stefanopoulou, ``Accelerated battery lifetime simulations using adaptive inter-cycle extrapolation algorithm,'' \emph{Journal of The Electrochemical Society}, vol. 168, no.~12, p. 120531, 2021.

\bibitem{11}
Z.~Bao, J.~Nie, H.~Lin, Z.~Li, K.~Gao, Z.~He, and M.~Gao, ``Dual-task learning for joint state-of-charge and state-of-energy estimation of lithium-ion battery in electric vehicle,'' \emph{IEEE Transactions on Transportation Electrification}, 2024.

\bibitem{12}
C.~Wang, R.~Wang, G.~Liu, Z.~Ji, W.~Shen, and Q.~Yu, ``Progressive degradation behavior and mechanism of lithium-ion batteries subjected to minor deformation damage,'' \emph{Journal of Energy Storage}, vol. 101, p. 113992, 2024.

\bibitem{13}
Y.~Wu, P.~Cong, and Y.~Wang, ``Charging load forecasting of electric vehicles based on vmd--ssa--svr,'' \emph{IEEE Transactions on Transportation Electrification}, vol.~10, no.~2, pp. 3349--3362, 2023.

\bibitem{r3_1}
R.~R. Richardson, M.~A. Osborne, and D.~A. Howey, ``Gaussian process regression for forecasting battery state of health,'' \emph{Journal of Power Sources}, vol. 357, pp. 209--219, 2017.

\bibitem{r3_5}
R.~Ibraheem, C.~Strange, and G.~Dos~Reis, ``Capacity and internal resistance of lithium-ion batteries: Full degradation curve prediction from voltage response at constant current at discharge,'' \emph{Journal of Power Sources}, vol. 556, p. 232477, 2023.

\bibitem{18}
W.~Duan, S.~Song, F.~Xiao, Y.~Chen, S.~Peng, and C.~Song, ``Battery soh estimation and rul prediction framework based on variable forgetting factor online sequential extreme learning machine and particle filter,'' \emph{Journal of Energy Storage}, vol.~65, p. 107322, 2023.

\bibitem{19}
Q.~Xu, M.~Wu, E.~Khoo, Z.~Chen, and X.~Li, ``A hybrid ensemble deep learning approach for early prediction of battery remaining useful life,'' \emph{IEEE/CAA Journal of Automatica Sinica}, vol.~10, no.~1, pp. 177--187, 2023.

\bibitem{nie1}
J.~Nie, Z.~He, Y.~Yang, Z.~Bao, M.~Gao, and J.~Zhang, ``Osp2b: One-stage point-to-box network for 3d siamese tracking,'' in \emph{Proceedings of the Thirty-Second International Joint Conference on Artificial Intelligence, {IJCAI-23}}, 8 2023, pp. 1285--1293.

\bibitem{nie2}
J.~Nie, A.~Xu, Z.~Bao, Z.~He, X.~Lv, and M.~Gao, ``Context matching-guided motion modeling for 3d point cloud object tracking,'' \emph{IEEE Transactions on Circuits and Systems for Video Technology}, 2024.

\bibitem{r1_4}
C.~Zhang, L.~Luo, Z.~Yang, B.~Du, Z.~Zhou, J.~Wu, and L.~Chen, ``Flexible method for estimating the state of health of lithium-ion batteries using partial charging segments,'' \emph{Energy}, vol. 295, p. 131009, 2024.

\bibitem{21}
M.~Reza, M.~Hannan, M.~Mansor, P.~J. Ker, S.~K. Tiong, and M.~Hossain, ``Gravitational search algorithm based lstm deep neural network for battery capacity and remaining useful life prediction with uncertainty,'' \emph{IEEE Transactions on Industry Applications}, 2024.

\bibitem{r1_1}
C.~Zhang, L.~Tu, Z.~Yang, B.~Du, Z.~Zhou, J.~Wu, and L.~Chen, ``A cmmog-based lithium-battery soh estimation method using multi-task learning framework,'' \emph{Journal of Energy Storage}, vol. 107, p. 114884, 2025.

\bibitem{r1_2}
C.~Zhang, Y.~Zhou, Z.~Zhou, S.~Chen, J.~Wu, and L.~Chen, ``State of health estimation of lithium-ion batteries based on multiphysics features and cnn-efc-bilstm,'' \emph{IEEE Sensors Journal}, 2024.

\bibitem{r3_3}
M.~Hassanaly, P.~J. Weddle, R.~N. King, S.~De, A.~Doostan, C.~R. Randall, E.~J. Dufek, A.~M. Colclasure, and K.~Smith, ``Pinn surrogate of li-ion battery models for parameter inference, part i: Implementation and multi-fidelity hierarchies for the single-particle model,'' \emph{Journal of Energy Storage}, vol.~98, p. 113103, 2024.

\bibitem{joule}
A.~Weng, E.~Dufek, and A.~Stefanopoulou, ``Battery passports for promoting electric vehicle resale and repurposing,'' \emph{Joule}, vol.~7, no.~5, pp. 837--842, 2023.

\bibitem{30}
H.~Liu, F.~Liu, X.~Fan, and D.~Huang, ``Polarized self-attention: Towards high-quality pixel-wise regression,'' \emph{arXiv preprint arXiv:2107.00782}, 2021.

\bibitem{31}
R.~Child, S.~Gray, A.~Radford, and I.~Sutskever, ``Generating long sequences with sparse transformers,'' \emph{arXiv preprint arXiv:1904.10509}, 2019.

\bibitem{NASA}
B.~Saha and K.~Goebel, ``“battery data set”, nasa prognostics data repository, nasa ames research center, moffett field, ca, usa,'' 2024.

\bibitem{CALCE}
W.~He, N.~Williard, M.~Osterman, and M.~Pecht, ``Prognostics of lithium-ion batteries based on dempster--shafer theory and the bayesian monte carlo method,'' \emph{Journal of Power Sources}, vol. 196, no.~23, pp. 10\,314--10\,321, 2011.

\bibitem{XJTU}
F.~Wang, Z.~Zhai, Z.~Zhao, Y.~Di, and X.~Chen, ``Physics-informed neural network for lithium-ion battery degradation stable modeling and prognosis,'' \emph{Nature Communications}, vol.~15, no.~1, p. 4332, 2024.

\bibitem{Oxford}
D.~Howey and C.~Birkl, ``Oxford battery degradation dataset 1,'' 2017.

\bibitem{MIT}
K.~A. Severson, P.~M. Attia, N.~Jin, N.~Perkins, B.~Jiang, Z.~Yang, M.~H. Chen, M.~Aykol, P.~K. Herring, D.~Fraggedakis \emph{et~al.}, ``Data-driven prediction of battery cycle life before capacity degradation,'' \emph{Nature Energy}, vol.~4, no.~5, pp. 383--391, 2019.

\bibitem{r3_6}
P.~Mohtat, S.~Lee, J.~B. Siegel, and A.~G. Stefanopoulou, ``Comparison of expansion and voltage differential indicators for battery capacity fade,'' \emph{Journal of Power Sources}, vol. 518, p. 230714, 2022.

\bibitem{R22}
S.~Pepe and F.~Ciucci, ``Long-range battery state-of-health and end-of-life prediction with neural networks and feature engineering,'' \emph{Applied Energy}, vol. 350, p. 121761, 2023.

\bibitem{c1}
Y.~Wei and D.~Wu, ``Prediction of state of health and remaining useful life of lithium-ion battery using graph convolutional network with dual attention mechanisms,'' \emph{Reliability Engineering \& System Safety}, vol. 230, p. 108947, 2023.

\bibitem{c3}
B.~Chen, Y.~Liu, and B.~Xiao, ``A novel hybrid neural network-based soh and rul estimation method for lithium-ion batteries,'' \emph{Journal of Energy Storage}, vol.~98, p. 113074, 2024.

\bibitem{c4}
Y.~Zhou, S.~Wang, Y.~Xie, X.~Shen, and C.~Fernandez, ``Remaining useful life prediction and state of health diagnosis for lithium-ion batteries based on improved grey wolf optimization algorithm-deep extreme learning machine algorithm,'' \emph{Energy}, vol. 285, p. 128761, 2023.

\bibitem{c_add_1}
Y.~Fan, Z.~Lin, F.~Wang, and J.~Zhang, ``A hybrid approach for lithium-ion battery remaining useful life prediction using signal decomposition and machine learning,'' \emph{Scientific Reports}, vol.~15, no.~1, p. 8161, 2025.

\bibitem{c_add_2}
T.~Zhu, W.~Wang, and M.~Yu, ``A novel hybrid scheme for remaining useful life prognostic based on secondary decomposition, bigru and error correction,'' \emph{Energy}, vol. 276, p. 127565, 2023.

\bibitem{c_add_4}
Y.~Li, X.~Qin, M.~Chai, H.~Wu, F.~Zhang, F.~Jiang, and C.~Wen, ``Soh evaluation and rul estimation of lithium-ion batteries based on mc-cnn-timesnet model,'' \emph{Reliability Engineering \& System Safety}, p. 111125, 2025.

\bibitem{c5}
X.~Li, C.~Yuan, Z.~Wang, J.~He, and S.~Yu, ``Lithium battery state-of-health estimation and remaining useful lifetime prediction based on non-parametric aging model and particle filter algorithm,'' \emph{Etransportation}, vol.~11, p. 100156, 2022.

\bibitem{c6}
Y.~Wei and D.~Wu, ``State of health and remaining useful life prediction of lithium-ion batteries with conditional graph convolutional network,'' \emph{Expert Systems with Applications}, vol. 238, p. 122041, 2024.

\bibitem{R21}
Y.~Sun, Q.~Diao, H.~Xu, X.~Tan, Y.~Fan, and L.~Wei, ``State of health estimation for lithium-ion batteries based on partial charging curve reconstruction,'' \emph{IEEE Transactions on Power Electronics}, 2024.

\end{thebibliography}

\begin{IEEEbiography}[{\includegraphics[width=1in,height=1.25in,clip,keepaspectratio]{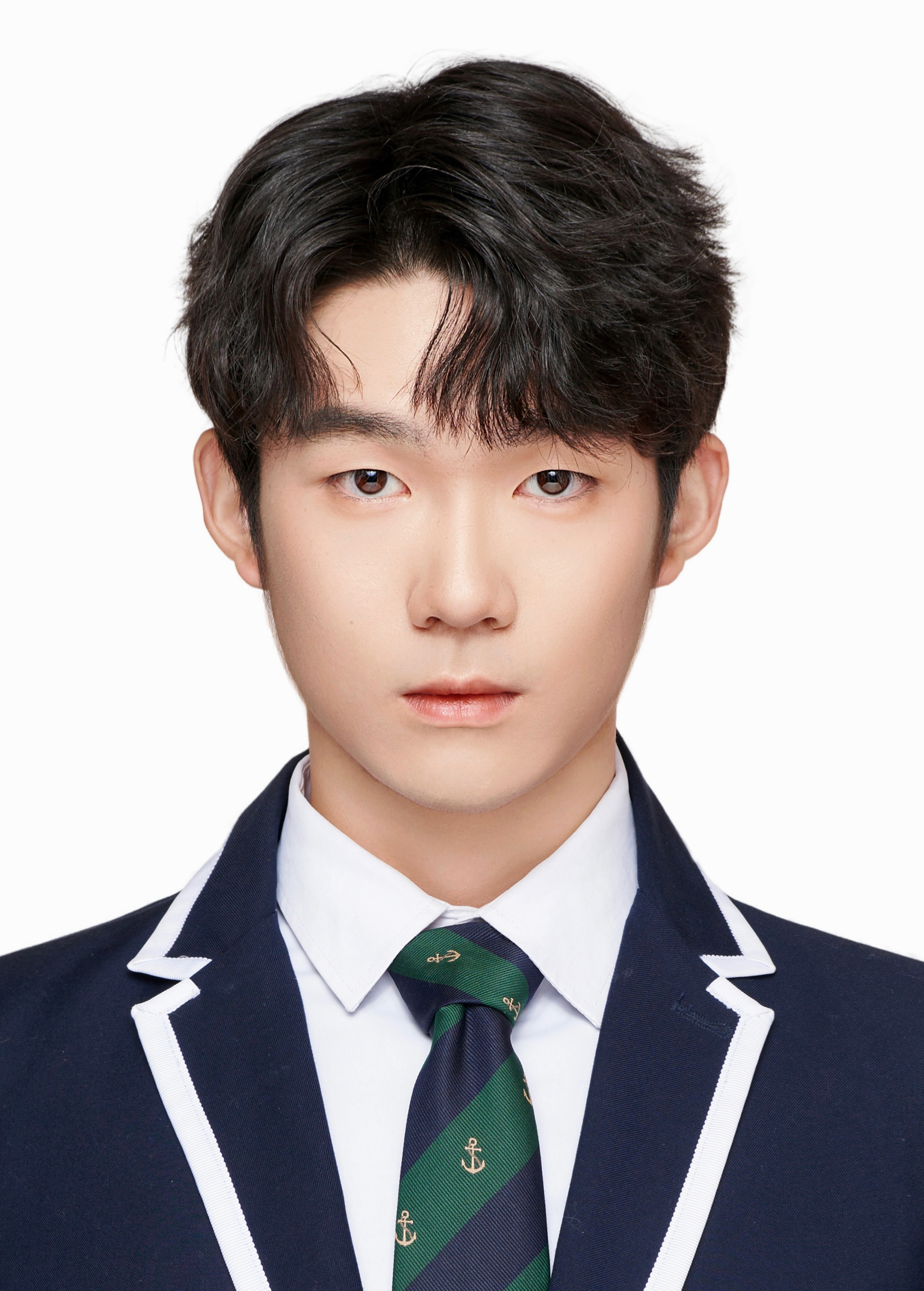}}]{Chenhan Wang}
  is currently pursuing the Bachelor of Engineering degree in the School of Electronic Information of Hangzhou Dianzi University. His current research focuses on the application of deep learning in battery management systems for electric vehicles, especially power battery health state prediction and remaining life prediction.
\end{IEEEbiography}

\begin{IEEEbiography}[{\includegraphics[width=1in,height=1.25in,clip,keepaspectratio]{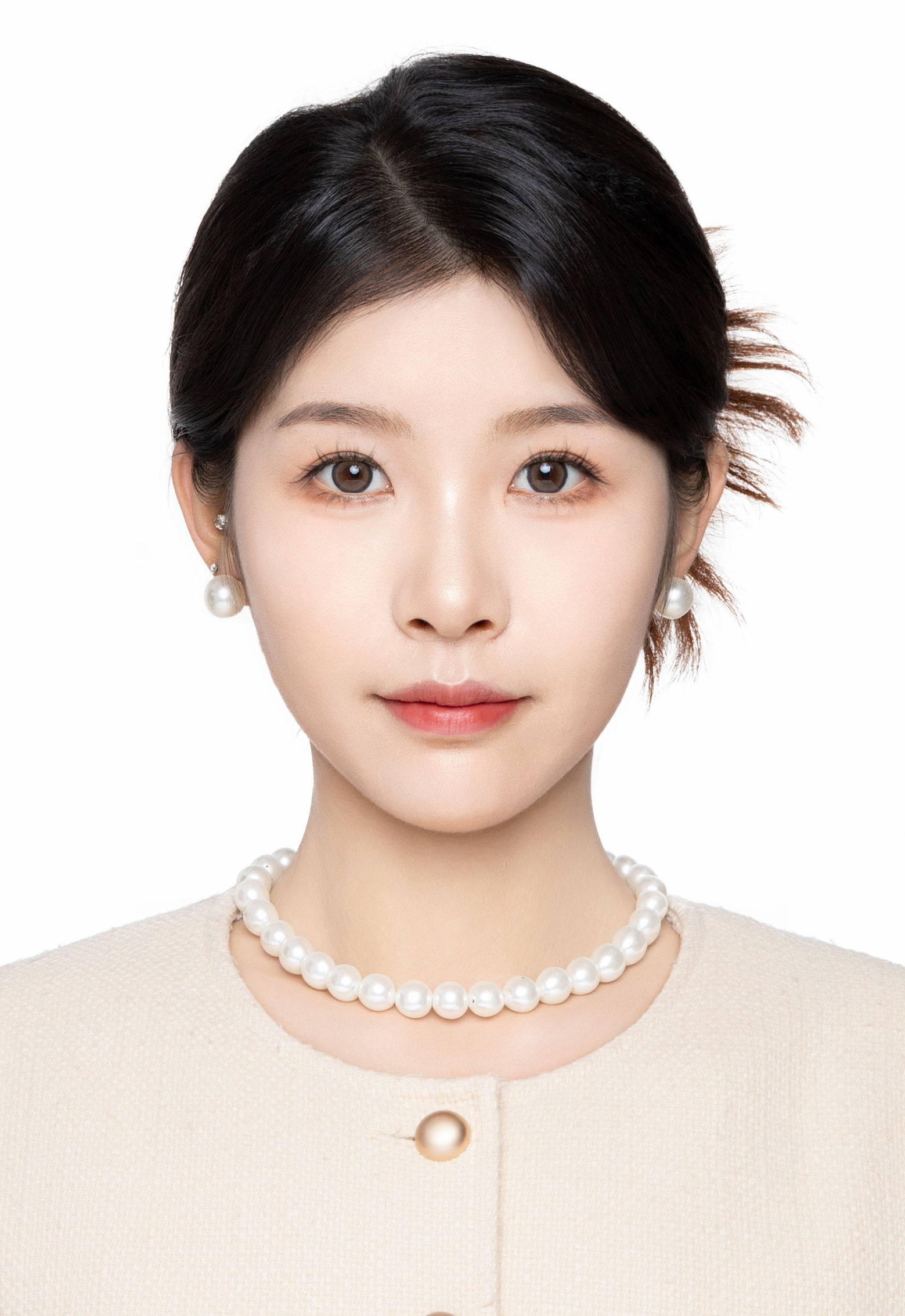}}]{Zhengyi Bao} 
   (Member, IEEE) received the B.S. degree from the School of Electronic Information at China Jiliang University, in 2021. She is currently working toward a Ph.D. degree at the School of Electronic and Information, Hangzhou Dianzi University. She has actively participated in several projects related to new energy electric vehicle battery management systems. Her research focuses on deep learning and neural networks for the state prediction of electric vehicle batteries, and has currently published several battery state prediction papers in journals such as \textsc{IEEE Trans. Transp. Electrif.} \textsc{and IEEE Trans. Veh. Technol.}
\end{IEEEbiography}

\begin{IEEEbiography}[{\includegraphics[width=1in,height=1.25in,clip,keepaspectratio]{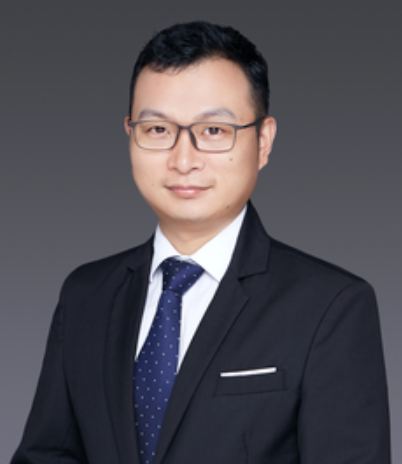}}]{Huipin Lin}
(Member, IEEE) received the Ph.D. degrees in Zhejiang University and currently working at Hangzhou Dianzi University. He has been engaged in research in battery management, new energy generation, power electronic converters, and Internet of Things technology for many years.
\end{IEEEbiography}

\begin{IEEEbiography}[{\includegraphics[width=1in,height=1.25in,clip,keepaspectratio]{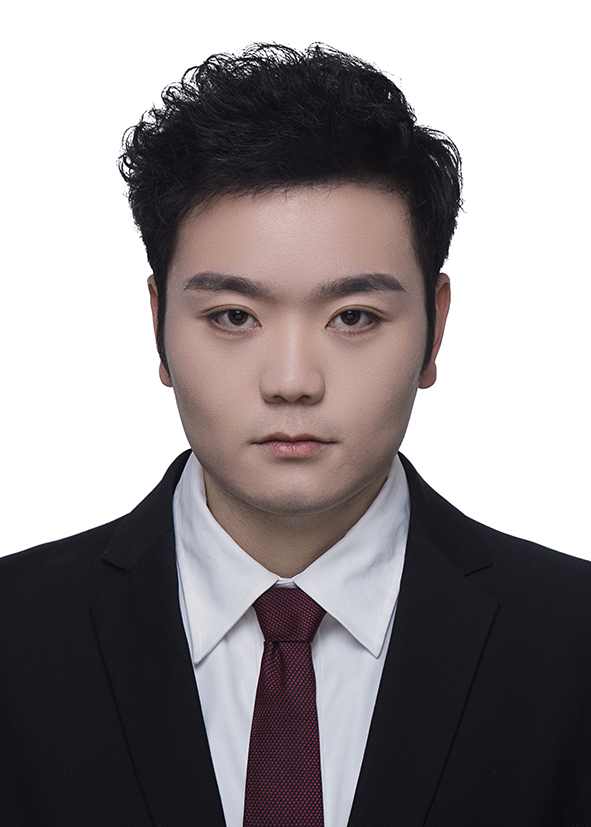}}]{Jiahao Nie}
   (Member, IEEE) received the Ph.D. degree at the School of Electronic and Information, Hangzhou Dianzi University in the year 2025. He is currently a assistant professor at the School of Information Technology and Artificial Intelligence, Zhejiang University of Finance and Economics. His research interests include deep learning and 2D/3D computer vision, especially on object detection and tracking. His current focus involves single object tracking and multi object tracking on Camera images and LiDAR point clouds. He has published more than 30 research papers in international conferences and journals, such as \textsc{IJCV}, \textsc{IEEE Tmm}, \textsc{IEEE Tcsvt},  \textsc{IEEE Tii}, \textsc{IEEE Tvt}, \textsc{IEEE Tte}, CVPR, ICLR, AAAI, IJCAI and ACM MM. 
\end{IEEEbiography}

\begin{IEEEbiography}[{\includegraphics[width=1in,height=1.25in,clip,keepaspectratio]{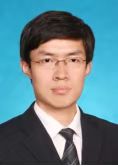}}]{Chunxiang Zhu} received his Ph.D. from Hangzhou Dianzi University and currently serves as a lecturer at China Jiliang University. His research interests include deep learning and lithium battery management systems, particularly in state estimation and early warning. His current research focuses on lithium battery health state estimation and remaining life prediction. He has published over 10 research papers in international conferences and journals such as IEEE Sensors Journal and Journal of Energy Storage.
\end{IEEEbiography}

\end{document}